\documentclass{article}

\usepackage{microtype}
\usepackage{graphicx}
\usepackage{subfigure}
\usepackage{booktabs} 

\usepackage{hyperref}
\usepackage{hyperref}       
\usepackage{url}            
\usepackage{booktabs}       
\usepackage{amsfonts}       
\usepackage{nicefrac}       
\usepackage{microtype}      
\usepackage{xcolor}         
\usepackage{graphicx}
\usepackage{amsfonts}
\usepackage{amsmath}
\usepackage{amssymb}
\usepackage{mathtools}
\usepackage{amsthm}
\usepackage{wrapfig}
\usepackage{hyperref}
\usepackage{bbding}

\usepackage[accepted]{icml2023}

\usepackage{amsmath}
\usepackage{amssymb}
\usepackage{mathtools}
\usepackage{amsthm}

\usepackage[capitalize,noabbrev]{cleveref}

\theoremstyle{plain}

\theoremstyle{definition}

\theoremstyle{remark}

\usepackage[textsize=tiny]{todonotes}

\PassOptionsToPackage{numbers}{natbib}
\setcitestyle{numbers}
\setcitestyle{square} 
\icmltitlerunning{Meta-Learning Parameterized Skills}

\begin{document}

\twocolumn[
\icmltitle{Meta-Learning Parameterized Skills}



\icmlsetsymbol{equal}{*}

\begin{icmlauthorlist}
\icmlauthor{Haotian Fu}{yyy}
\icmlauthor{Shangqun Yu}{zzz}
\icmlauthor{Saket Tiwari}{yyy}
\icmlauthor{Michael Littman}{yyy}
\icmlauthor{George Konidaris}{yyy}
\end{icmlauthorlist}

\icmlaffiliation{yyy}{Department of Computer Science, Brown University}
\icmlaffiliation{zzz}{The University of Massachusetts Amherst}
\icmlcorrespondingauthor{Haotian Fu}{hfu7@cs.brown.edu}


\vskip 0.3in
]


\printAffiliationsAndNotice{}


\begin{abstract}
We propose a novel parameterized skill-learning algorithm that aims to learn transferable parameterized skills and synthesize them into a new action space that supports efficient learning in long-horizon tasks. We propose to leverage off-policy Meta-RL combined with a trajectory-centric smoothness term to learn a set of parameterized skills. Our agent can use these learned skills to construct a three-level hierarchical framework that models a Temporally-extended Parameterized Action Markov Decision Process. We empirically demonstrate that the proposed algorithms enable an agent to solve a set of difficult long-horizon (obstacle-course and robot manipulation) tasks.
\end{abstract}

\section{Introduction}
\begin{figure*}[htbp]
\centering
    \includegraphics[width=0.9\linewidth]{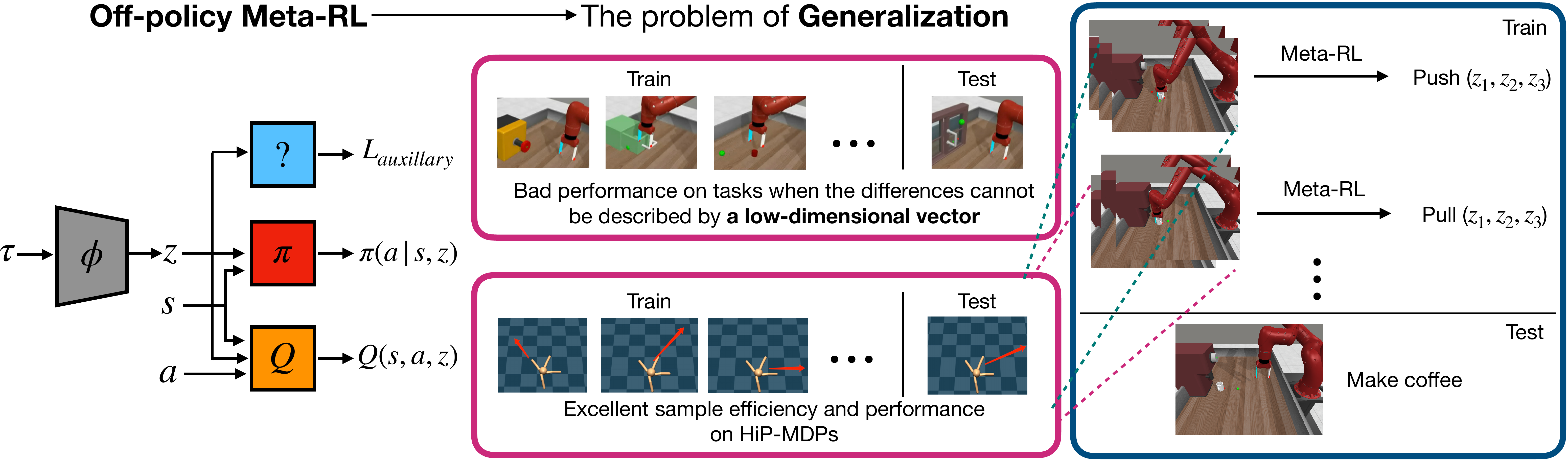}

    \caption{Left: Off-policy Meta-RL. The meta-policy $\pi$ takes in the state as well as a latent vector $z$ as input. On a new task, the context encoder $\phi$ will try to find the latent vector corresponding to the current task from a few trajectories $\tau$. Mid: Off-policy Meta-RL in two different scenarios. Right: Leveraging off-policy Meta-RL to learn parameterized skills.} 
    \label{fig:opml}
\end{figure*}

To improve Reinforcement Learning (RL)'s generalization to novel tasks, meta-Reinforcement Learning (meta-RL) learns a meta-policy from a large number of tasks that aims to quickly adapt to a new task within the same distribution. Off-policy meta-RL methods~\citep{DBLP:conf/icml/RakellyZFLQ19, 
DBLP:conf/icml/LeeSLLS20, DBLP:conf/aaai/FuTHCFLL21,  DBLP:conf/nips/DorfmanST21} normally train a context-encoder that takes in a few collected trajectories/transitions on a new task as input and output latent parameters that function as a descriptor of the current task. That descriptor is fed into the policy as an additional input to generate actions. Compared to On-policy meta-RL methods~\citep{Finn2017ModelAgnosticMF, DBLP:journals/corr/WangKTSLMBKB16, Zintgraf2020VariBADAV},  off-policy methods generally have much higher sample efficiency and better or comparable overall performance~\citep{DBLP:conf/icml/NiES22, Zintgraf2020VariBADAV, DBLP:conf/icml/RakellyZFLQ19} on tasks {\bf whose differences vary smoothly and can be described by a single vector} (e.g., tasks change between different goal velocity for a half-cheetah)---a setting also known as Hidden-parameter MDPs (HiP-MDPs)~\citep{DoshiVelez2016HiddenPM, DBLP:journals/corr/KillianDKD17, fu2023performance}. 
However, for tasks with more diverse variations (e.g., tasks change between pull the mug, press the button, open the door, etc., see Figure~\ref{fig:opml}), off-policy methods fail to generalize well compared to on-policy methods and methods based on fine-tuning~\citep{DBLP:conf/corl/YuQHJHFL19, DBLP:journals/corr/abs-2206-03271}, even given a much larger number of adaptation steps. This makes off-policy methods hard to to apply to realistic problems despite their superiority on HiP-MDP environments.

However, fast adaptation of an entire policy to a new task is not the only possible form of generalization that we may want  RL agents to display. Another approach is learning  reusable high-level skills \citep{Sutton1999BetweenMA}, which enable an agent to explore efficiently and solve hard long-horizon tasks using hierachical methods. 
In realistic tasks, we want skills that are flexible---able to be efficiently adapted to many different situations. For example, a skill that opens a door should be adjustable to many different types of doors and handles, from office doors to microwave doors.
The most flexible skills are \emph{parametrized}: discrete skills augmented with continuous parameters that adjust their behavior, thereby making them more likely to be reusable in new tasks because they are flexible enough to be applied in diverse situations. Finding the appropriate parametrization of a skill abtracted from the primitive action space in such settings is still an open  question.

We propose that the problem of learning parameterized skills is very similar to the HiP-MDP setting, in which off-policy meta-RL methods successfully generalize. 
Specifically, by leveraging Off-policy Meta-RL, we propose to learn \emph{parameterized skills}~\citep{DBLP:conf/icml/SilvaKB12}---both the skills themselves and the parameter space---as well as a high-level control policy that will use the learned parameterized skills as the new action space and perform on new tasks. Our contributions are: 1. We propose a novel three-level hierarchical RL framework combining off-policy Meta-RL and Parameterized Action MDP algorithms (MLPS + HPS) to model Temporally-extend PAMDP problems, which can be used to solve long-horizon tasks. 2. For low-level policy learning, we propose a novel trajectory-centric smoothness training objective for learning parameterized skills capable of expressing diverse behaviors with a smooth parameter space. 3. For high-level and mid-level policy learning, we propose a novel hierarchical actor-critic algorithm that, given the learned parameterized action space, exhibits better performance compared to previous PAMDP algorithms. 4. Using the proposed algorithm, we are able to solve a set of difficult long-horizon ant obstacle course tasks, as well as long-horizon robotic manipulation tasks. 5. We demonstrate the importance of smoothness for a learned parameterized action space and the effectiveness of the different components of our algorithm independently.\footnote{A video of the learned policy can be found at~\url{https://youtu.be/Ux2s_BbED9Q}. Our code is available at \url{https://github.com/Minusadd/Meta-learning-parameterized-skills}.}

\section{Background}

A Parameterized Action Markov Decision Process (PAMDP)~\citep{DBLP:conf/aaai/MassonRK16} is defined by the tuple $\{S, H, T, R, \gamma\}$, where the parameterized action space $H$ can be defined as: $H = \{(k, z_{k})|z_{k} \in Z_{k} ~\text{for all}~ k \in \{1, \cdots, K\} \}$, where $z_{k}$ is the corresponding continuous parameter set for each discrete action $k$. Here, $z_{k}$ is the continuous parameter corresponding to $k$, and $K$ is the total number of discrete actions. At each step, the agent must select both a discrete action $k$ and a continuous parameter $z_k$. Thus, we have the dynamic transition function $T(s'|s,k,z_{k})$ and the reward function $R(r|s,k,z_{k})$. 
A practical example is a football game, where the player needs to choose between kick the ball or move to some position (discrete), as well as the direction the player wants to kick the ball to or the specific position the player wants to move to (continuous). Most previous work assumes the primitive action space is parameterized, or a set of predefined parameterized skills are given. Our work makes an attempt to learn/synthesize the parameterized action space from scratch.

HiP-MDPs model the variations in the transition dynamics and reward functions by assigning each task a hidden parameter $\theta$, drawn from the distribution $P_{\Omega}$. 
The agent neither observes $\theta$ nor has access to the distribution $P_\Omega$ that generates the task family. 
For a given task, parameterized by $\theta \in \Theta$, the stochastic dynamics are given by $T(s'|s,a;\theta)$ and the deterministic reward function by $R(s,a;\theta)$. A commonly-used meta-RL benchmark creates a set of tasks by changing the environmental parameters (e.g. mass, damping)~\citep{DBLP:conf/icml/LeeSLLS20, DBLP:conf/icml/RaileanuGSF20, DBLP:conf/nips/FuYL022} or reward functions (e.g. target position, target velocity)~\citep{DBLP:conf/icml/RakellyZFLQ19, Zintgraf2020VariBADAV} of Mujoco-simulated robots. 

Off-policy Meta-RL (OPML), shown in Figure~\ref{fig:opml}, learns a meta-policy $\pi(a|s,z)$ that is shared across all the tasks from the same distribution, as well as a context encoder $\phi(z|\tau)$ that maps collected transitions $\tau = \{s_1, a_1, r_1, s_2, \cdots, s_n\}$ to a task encoding $z$. The learned task encoding should indicate how the underlying hidden parameter $\theta$ changes the optimal policy in the HiP-MDP. When facing a new task, the agent  interacts with the environment for a few episodes and inputs the resulting trajectories into the context encoder, from which it can  infer  the corresponding latent parameter to the policy. To train the context encoder, previous work uses the critic loss~\citep{DBLP:conf/icml/RakellyZFLQ19}, or some auxiliary loss like the dynamics prediction~\citep{DBLP:conf/icml/LeeSLLS20, DBLP:conf/nips/DorfmanST21, DBLP:conf/l4dc/SodhaniMP022} or contrastive loss~\citep{DBLP:conf/aaai/FuTHCFLL21}.

\section{Meta-Learning Parameterized Skills}

\begin{figure}[htbp]
\centering
    \includegraphics[width=0.75\linewidth]{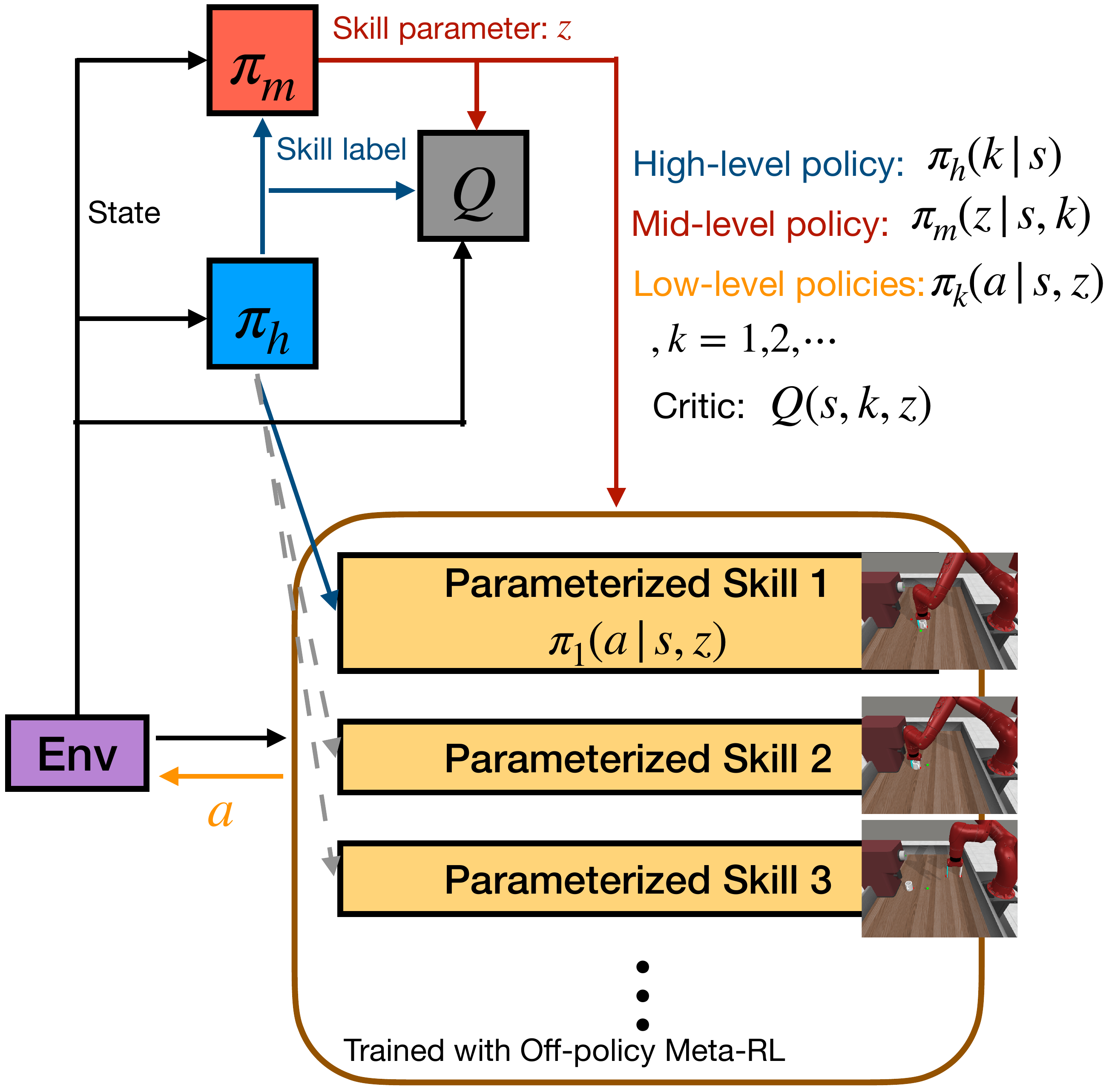}

    \caption{Meta-learning parameterized skills: a three-level hierarchical framework modeling a TPAMDP. The learned parameterized skills are treated as a parameterized action space for the high- and mid-level policies, while each of the skills is actually a temporally abstraction of the low-level policy on the primitive action space.} 
    \label{fig:alg}
\end{figure}

In general, we want the agent to learn a set of parameterized skills suitable to be used as the parameterized action space in a PAMDP, for which the agent will in turn learn a high-level control policy to solve new tasks. We show the overall three-level hierarchical framework of our proposed algorithm in Figure~\ref{fig:alg}. We use this hierarchical framework to model a Temporally-extended PAMDP (TPAMDP). At the beginning of one episode, the agent receives a state from the environment. The state will be passed to the high-level policy $\pi_h$ first, which will output the discrete skill label $k$. Then the skill label and the state will be fed into the mid-level policy $\pi_m$, which will output the skill parameter $z$ corresponding to skill $k$. The agent will then choose the low-level policy $\pi_k$ corresponding to the skill label $k$ as the current executing policy, which will take the state and skill parameter $z$ as input and output primitive actions. The low-level policy $\pi_k$ will interact with the environment for $T$ steps, after which the high-level policy will receive a new state and carry out the same process to choose the skill label and the corresponding parameters again. Overall, for a TPAMDP, we have a high-level policy and a mid-level policy that solve a new task by mapping the states to parameterized skill pairs $(k, z)$---learning in the high-level temporally extended parameterized action space. Each discrete skill label $k$ corresponds to a low-level skill-conditioned policy network $\pi_k(a|s, z)$, which takes the continuous skill parameter $z$ as an additional input. As the low-level policies are fixed, they can be treated as part of the environment during the training of high and mid-level policies. In Sec.~\ref{met:low}, we introduce how our agent learn the low-level policy (MLPS). In Sec.~\ref{sec32}, we explain how our agent learns the high-level and mid-level policies (HPS).

\subsection{Off-policy Meta-RL for Parameterized Skills }
\label{met:low}
We first address how to learn the continuous parameters associated with each discrete action (skill category) to cover policies with similar and smoothly changing behaviors. To this end, 
 we model a parameterized skill as a HiP-MDP, meaning the agent is given a set of tasks that share similar reward/dynamics structure. 
By modeling the parameterized skill as a HiP-MDP, the task set that we train our agent on has an underlying and potentially smoothly-varying hidden parameter that controls the distinct features of each task. 

Ideally, we want the agent to learn a policy that is able to solve the HiP-MDP---a robust skill-conditioned policy, and also learn a continuous representation $z$ that smoothly approximates how the hidden parameters $\theta$ affect the agent's optimal policy on each task. Using off-policy Meta-RL, it is straightforward way to let the agent learn a skill-conditioned policy that additionally takes the continuous representation $z$ as a input: $\pi: S\times Z \rightarrow A$. Then, given different values of $z$, the policy will output actions that can solve different tasks. By leveraging the high sample efficiencyof Off-policy Meta-RL, we can get a high-performing skill-conditioned policy quickly. We let the agent learn $K$ different skill-conditioned policies, which will be fixed as the low-level policies during the following higher-level policies' training.

\newcommand{\Value}{\mathrm{Value}}

For practical implementation, we adopt the framework of a recent off-policy Meta-RL algorithm, PEARL~\citep{DBLP:conf/icml/RakellyZFLQ19}, and train a context encoder that aims to put the collected trajectories into a latent representation, along with an actor and a critic network that both take in the latent representation as an additional input. In particular, we train a context encoder network $\phi: \tau \rightarrow z$ that generates latent representation $z$ using historical transitions. Then, the generated $z$ can be viewed as part of the state and can help the decision-making process as input to the actor network $\pi(a|s, z)$ and critic network $Q(s,a,z)$ as in PEARL. We provide more detailed algorithm and implementation information in Appendix~\ref{app:algo}.

\begin{figure}[htbp]
\centering
    \includegraphics[width=0.96\linewidth]{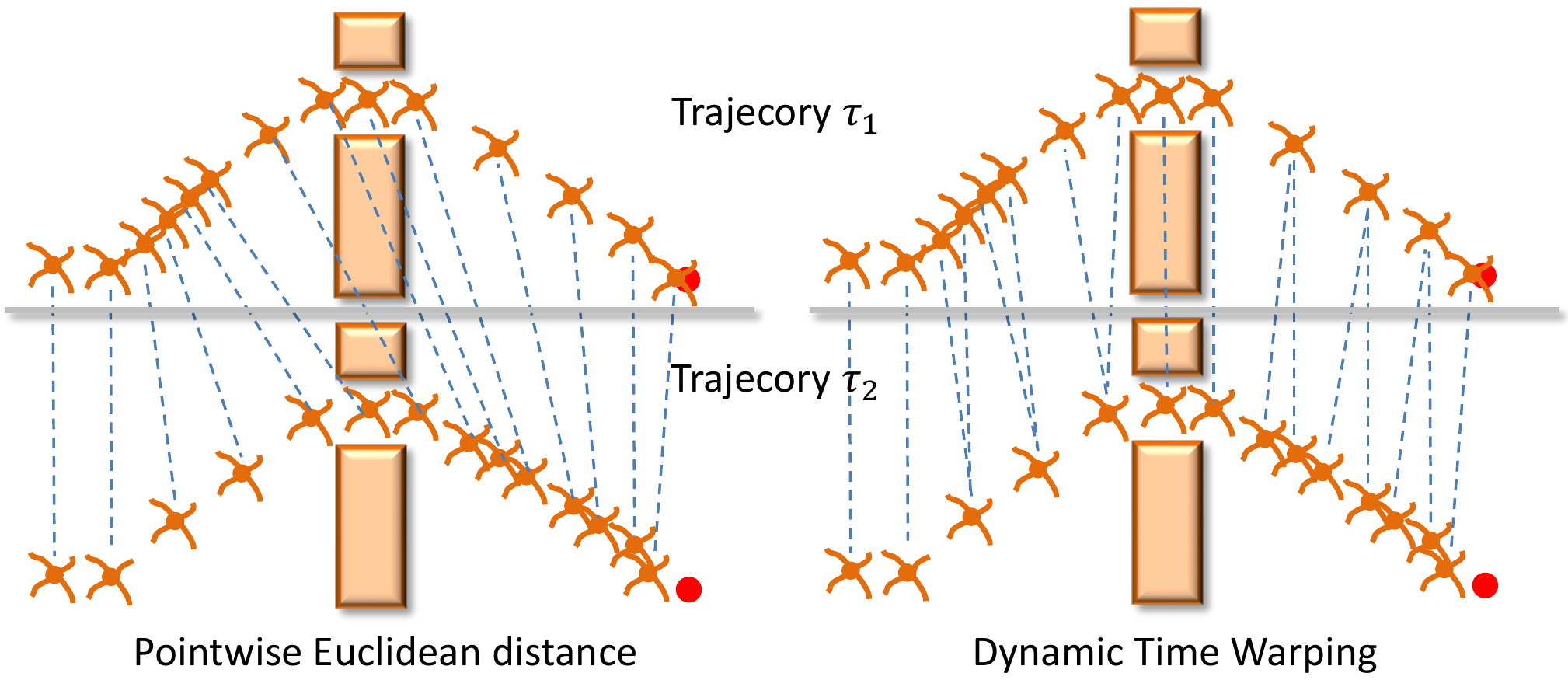}
    \caption{Trajectories' Dynamic Time Warping distance compared with Pointwise Euclidean distance. Trajectory $\tau_{1}$ and $\tau_{2}$ are sampled from the task. Using Dynamic Time Warping to compute the distance (right) reveals they are quite close. However, unwarped pointwise Euclidean distance (left) ends up with the erroneous conclusion that the trajectories are very different.} 
    \label{fig:dtw}
\end{figure}

{\bf Trajectory-Centric Smoothness} In the parameterized skill-learning setting, besides the goal of learning a policy that performs well in all tasks, we also want that the continuous representation $z$ which the policy is conditioned on is able to smoothly varying the agent's behaviors so that we can get a new smooth action space for this skill type and is reusable in other contexts. To achieve this goal, we propose the trajectory-centric smoothness training objectives for training the context encoder network. Note that instead of focusing on the difference between single transitions~\citep{DBLP:conf/iclr/EysenbachGIL19}, we propose that parameterized skill learning should focus more on the overall difference between different trajectories. The learned representation of the skill should be able to encode the distinguishable features of the trajectories into its continuous parameters.  Previous work shows the importance of smoothness in state representation learning~\citep{DBLP:conf/icml/GeladaKBNB19, DBLP:conf/iclr/0001MCGL21, DBLP:conf/nips/AllenPGK21}. Our case can be seen as policy representation learning, as we will use the learned representation space as the new action space, better smoothness intuitively will help the agent learn to identify the values of the continuous parameters for a new task more quickly. In Section~\ref{exp:smooth}, we empirically show how the smoothness of the learned skill parameter space will affect the overall performance of the algorithm. We propose that the agent's behavior under the skill-conditioned policy should change proportionally to the change of the continuous parameters' value. We hope to implicitly encode the semantic meanings of the underlying hidden parameters into our latent skill representation, thus improve the smoothness of the latent skill embedding space. Therefore we add another learning objective that aims to embed intermediate features of the state trajectories into the latent representation. Our main intuition is that {\bf the distance of different skills in the latent space should be proportional to the distance between their trajectories}. Specifically, suppose we sample two batches of trajectories $\tau_{1}$ and $\tau_{2}$ from two different tasks. Then, we write the smoothness term as:
\newcommand{\Smoothness}{\mathrm{Smoothness}}
\newcommand{\DTW}{\mathrm{DTW}}
\begin{equation}
\label{eqn:dtw}
    L_{\Smoothness} := MSE[||\phi(\tau_{1}) - \phi(\tau_{2})||_{2}- \kappa \DTW(\tau_{1},\tau_{2})],
\end{equation}
where $\DTW$ stands for Dynamic Time Warping~\citep{Bellman1959OnAC, Mller2008DynamicTW}, and $\kappa$ controls the scale of the DTW distance. Instead of directly computing the Euclidean distance between two state trajectories, we use Dynamic Time Warping to align the trajectories before measuring the distance. The idea is illustrated in Figure~\ref{fig:dtw}. Even from the exact same state and using the same policy, the pointwise Euclidean distance between two trajectories can be large as there exists uncertainty in both the environmental dynamics and the output actions from the policy. Thus, we use a more reasonable metric that compares the overall ``shape'' of the two trajectories, which is more consistent with our goal of extracting the overall features of the trajectory instead of focusing on specific transitions. By minimizing the smoothness term, we obtain skill embeddings that correspond to the dynamic time warping distance of trajectories.

\subsection{Hierarchical actor-critic with Parameterized Skills}
\label{sec32}
Then, given a set of low-level parameterized skills, the remaining question is how to efficiently learn high-level and mid-level control policies of our three-level hierarchical model in this temporally-extended PAMDP. As the low-level policies are fixed, the interaction between these higher-level policies and the environment is very close to a standard PAMDP. Thus a straightforward way is to directly apply PAMDP algorithms. HyAR~\citep{Li2021HyARAD} is a recently proposed algorithm that constructs a latent embedding space to model the dependency between discrete actions and continuous parameters. 
The discrete action along with the continuous parameters are mapped into a single latent action space, for which a policy is learned. However, learning directly in this latent embedding space means that the quality of exploration highly depends on whether the embedding space is learned properly. This problem becomes more severe as our parameterized action space are  learned from data and can be quite noisy. That is, given the same state and the same parameterized skills, the distribution of the next state might have large uncertainty because executing each skill involves a large number of steps' interaction with the environment, of which the resulting trajectories could be quite noisy. Thus, learning to embed this generated action space further into some latent space may magnify the uncertainty of transitions. Another straightforward but effective approach is P-DQN~\citep{Bester2019MultiPassQF, Xiong2018ParametrizedDQ}. The P-DQN agent  maintains a separate policy network for each discrete action $k$ to output the corresponding continuous parameters, and then feed all these parameters from different discrete actions into the critic network. This makes computation highly expensive as it always has to compute all the continuous parameters for each discrete action, and is magnified when the number of discrete actions are large. In our case, to enable structured exploration at both discrete action and continuous parameter level, we propose to directly model the dependency of the discrete and continuous part of the parameterized action with two consecutive policy networks: for each decision-making step, we first choose the discrete action, then choose the continuous parameters conditioned on both the state and discrete action, which is in consistent with human's decision making process~\citep{Parr2018TheDA}.

Concretely, as shown in Figure~\ref{fig:alg}, we decompose the policy of parameterized actions as:
\[
    \pi(k, z_{k}|s) = \pi_{\theta_{c}}(z_{k}|s,k) \pi_{\theta_{d}}(k|s),
\]
 where the policy network for discrete part of the action takes in state $s$ and is parameterized by $\theta_d$, the policy network for the continuous parameter $z_k$ takes in state and the discrete action $k$ output from $\pi_{\theta_d}$ and is parameterized by $\theta_c$. Compared with P-DQN, we only need to compute the continuous parameters for the discrete action we chose and thus avoid the redundancy problem.  

We update the policy using actor-critic framework with the maximum entropy learning objective for reinforcement learning~\citep{Ziebart2008MaximumEI, Haarnoja2018SoftAO}.  Maximum entropy RL greatly improve the exploration especially in the face of estimation error. It functions by maximizing the entropy of the policy as well as the expected return. This particularly fits our framework as the parameterized action space is learned and can be quite noisy. Further, exploration with different rates at different time periods of training is important in the long-horizon tasks as we explained in introduction. Concretely, we update the critic network $Q_{\psi}(s,k,z_k)$ according to:
\[
    L_{critic} = \mathbb{E}_{(s,k,z_{k}, r, s')\sim B}[Q_{\psi}(s,k,z_k) - (r + \gamma V(s))]^2,
\]
 where $B$ denotes the replay buffer, $V(s)$ denotes the value network. We update the policy (actor) networks according to:
 \begin{equation*}
 \begin{aligned}
       &L_{actor} = \\ &\mathbb{E}_{s\sim B, k\sim GS[\pi_{\theta_d}(s)]}\Big [D_{KL}\Big(\pi_{\theta_c}(z_{k}|s,k)\Big|\Big|\frac{\exp(Q_{\psi}(s,k,z_k))}{W_{\psi}(s)} \Big) \Big],  \end{aligned}
 \end{equation*}

 where $W_{\psi}(s)$ is the partition function that normalizes the distribution, $GS$ denotes the \textbf{gumbel-softmax} distribution~\citep{Jang2017CategoricalRW}. That is, to enable structured exploration at different levels of the action execution phase, we use the maximum entropy training objective to augment exploration for the policy of continuous parameters $\pi_{\theta_c}$, while we use gumbel-softmax technique to sample the discrete action to further augment the exploration for the policy of discrete action $\pi_{\theta_d}$. Compared to $\epsilon$-greedy exploration strategy, gumbel-softmax further augments structured exploration by sampling from the categorical distribution. It enables computing gradients for parameters of $\pi_{\theta_d}$, of which the outputs are discrete, by leveraging the reparameterization trick~\citep{Kingma2014AutoEncodingVB}. We use gumbel-softmax to sample from the discrete policy network when interacting with the environment during training and also when updating the network. The latter one uses a smaller value of temperature $\tau$ (controls the exploration rate) to make the updating process smoother following the intuition in~\citep{Fujimoto2018AddressingFA}.

Note that HHQN~\citep{Fu2019DeepMR} which focuses on multi-agent problem domain also uses a similar consecutive policy networks structure. However, they use two different Q networks to approximate the value of discrete and continuous policy which may cause high-level non-stationary problem~\citep{Li2021HyARAD}, i.e. when sampling a transition from the replay buffer, the same discrete action may not lead to the same reward and next state as the continuous parameter can be different from the moment it was chosen. Thus, computing the Q-value of a discrete action without considering the continuous parameter can be quite noisy. We avoid this problem as HPS has only one critic network that measures the value of the hybrid action pair as a whole.

\begin{figure}[htbp]
\centering
    \includegraphics[width=0.6\linewidth]{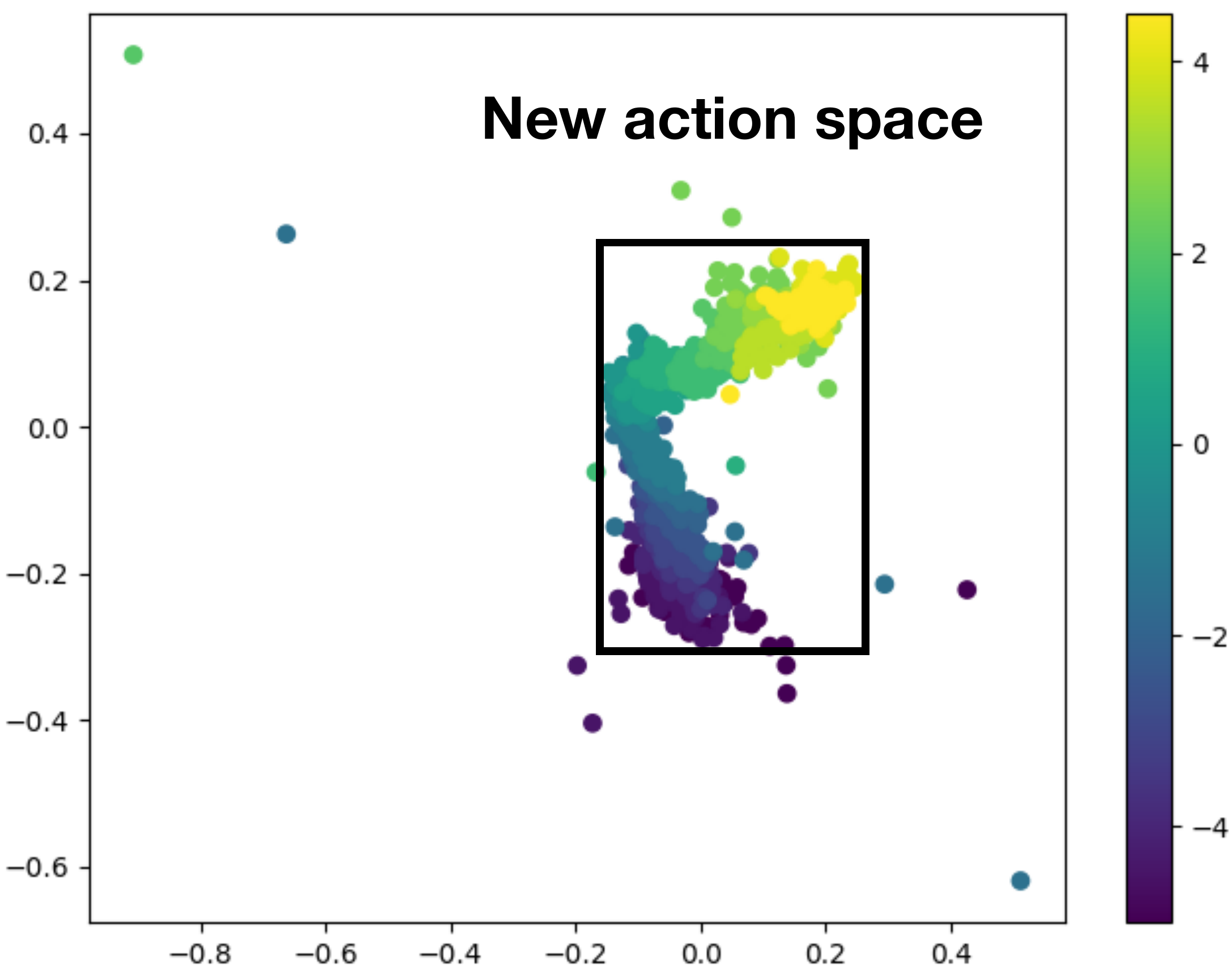}
    \caption{Visualization of the learned representation space of one skill. All the data points are from the same skill label $k$ but with different values of skill parameter $z$. } 
    \label{fig:actbound}
\end{figure}
{\bf New action space constraint} For practical implementation, as we are using the learned skills as a new action space for the higher-level policies, we also need to find and add constraints to the values of the action space that the mid-level policy can choose from. For each category of skills, we first run the standard meta-test process across across all the available training tasks for multiple times and collect the value of skill parameters $z$. As shown in Figure~\ref{fig:actbound}, most of the learned representations are close to each other in the latent space, but there are always outliers that are far away from the main cluster. If we set the bounds of the value of the action space to contain all these data points, the blank area between the outliers and the main cluster, also called "unreliable areas",  may deteriorate the higher-level policies, as shown in~\citep{DBLP:conf/corl/ZhouBH20, Notin2021ImprovingBO, Li2021HyARAD}. Thus, in practice, we rescale each dimension of the learned action space to a new bounded area by calculating the $t \%$ central range over the values of the collected data points, where $t \in [90, 100)$. 
\section{Experiments}
\label{exp}

\begin{figure*}[htbp]
\centering
    \includegraphics[width=0.94\linewidth]{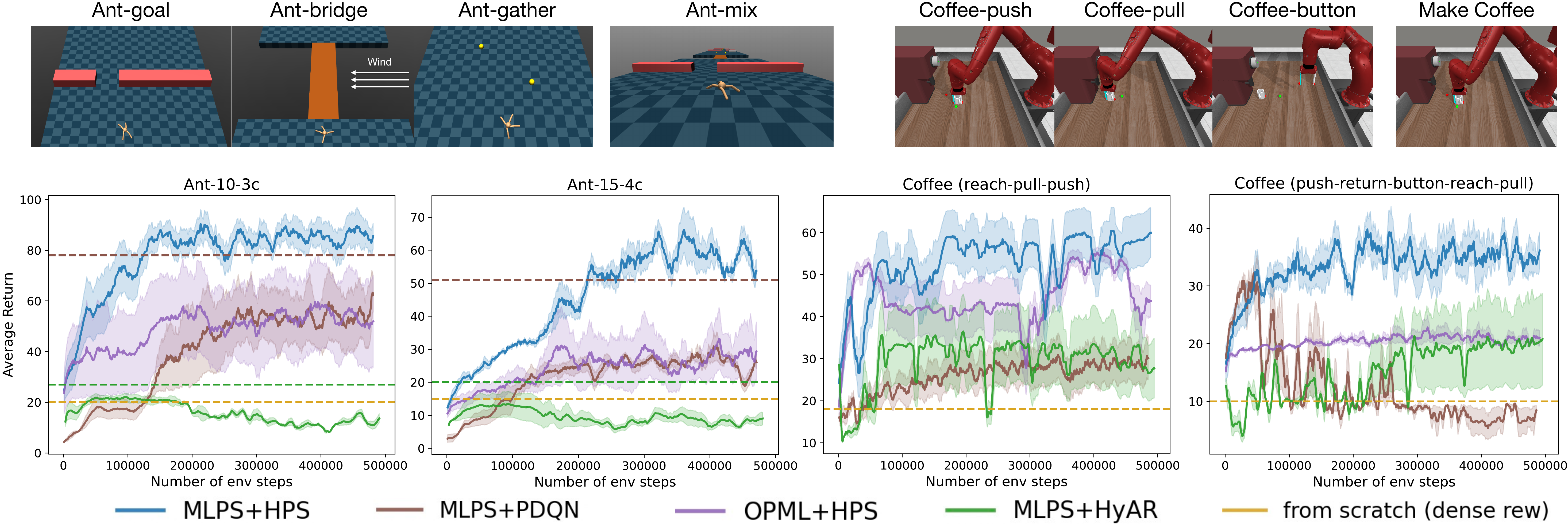}

    \caption{First row: The environments we used for Parameterized skill learning experiments. Second row: Comparison results of our method MLPS + HPS against other baselines in four scenarios . The horizontal axis denotes the number of ``env'' steps the {\bf high-level} agent takes instead of the original environment steps. Dashed lines correspond to the {\bf maximum} average return achieved by MLPS+PDQN and MLPS+HyAR after 1e6 ``env'' steps, as well as the maximum average return achieved by SAC learning from scratch using {\bf dense reward}.} 
    \label{fig:res1}
\end{figure*}
As shown in Figure~\ref{fig:res1} first row, we evaluate our algorithm on a Ant obstacle course domain built on OpenAI gym~\citep{1606.01540} and a robotic manipulation domain from MetaWorld~\citep{DBLP:conf/corl/YuQHJHFL19}. Long-horizon tasks at the level of primitive actions are highly difficult (see Appendix~\ref{app:longhor}) and can be reduced to very short-horizon tasks with the help of skills.

Ant-mix (obstacle course) have 10/15 consecutive barriers (denoted as 10-3c and 15-4c respectively in the plots) sampled from 4 categories of tasks: Ant-Goal, Ant-Bridge, Ant-Gather and Ant-box. Ant-Goal requires the agent to walk past a doorway at a position unknown and unseen to the agent, and reach the goal on the other side. Ant-Bridge requires the agent to walk across a bridge with cross wind. The speed of the wind is unknown to the agent. Ant-Gather requires the agent to gather two coins along its way to the goal position. The positions of the two coins are unknown to the agent. The agent succeeds after it reaches the goal position, which is fixed across all the tasks. The input states consist of the ant's position and other proprioceptive state, i.e., the angle/velocity of different joints. In these three tasks, the positions of the coins, the position of the doorway, and the wind speed are the corresponding hidden parameters in their MDPs, and the values of them are all sampled independently from a uniform distribution. 

The \emph{Make Coffee} task requires the robot arm to push the mug under the coffee machine, press the button, return to the original position, reach the mug and pull the mug to the target position. We train the agent to learn three parameterized skills: Coffee-push, Coffee-pull and Coffee-button as well as two discrete skill: reach and return. The input states are the proprioceptive state of the robot arm, as well as the position of the mug. For the high-level and mid-level policies, we also include the label of the current subtask we want to agent to do (e.g, push, pull, etc.). Otherwise, the environment would be non-stationary (i.e., same state-action pair but different reward.) For training the three parameterized skills, the target position we want to push/pull the mug to, and the position of the button are the corresponding hidden parameters in their MDPs, and the values of them are all sampled independently from a uniform distribution. 

We run MLPS as well as standard Off-Policy Meta-RL (OPML) on each of the HiP-MDPs and get the skills $\{Goal(x_1, x_2), Bridge(x_1, x_2), Gather(x_1, x_2), Box()\}$ for Ant-mix and $\{Push(x_1, x_2, x_3, x_4),Reach(), Return()\\ Button(x_1, x_2, x_3, x_4), Pull(x_1, x_2, x_3, x_4)\}$ for Make-coffee. For each random seed of training, we sampled the order of the subtasks (Make-coffee) as well as the hidden parameters of each subtask at the beginning of the experiment and fixed them for the rest of training and evaluation. We then used the  parameterized skills learned in previous section as the new parameterized action space, and let HPS learn a solution policy for it.  
We give the agent {\bf sparse staged reward}: a positive reward is received only when the ant has completed a subtask or it reaches the final goal, otherwise, the reward is 0. More environmental details and experiments can be found in Appendix~\ref{app:env}$\sim$\ref{hrlbaseline}. 

\begin{figure*}[htbp]
\centering
    \includegraphics[width=0.7\linewidth]{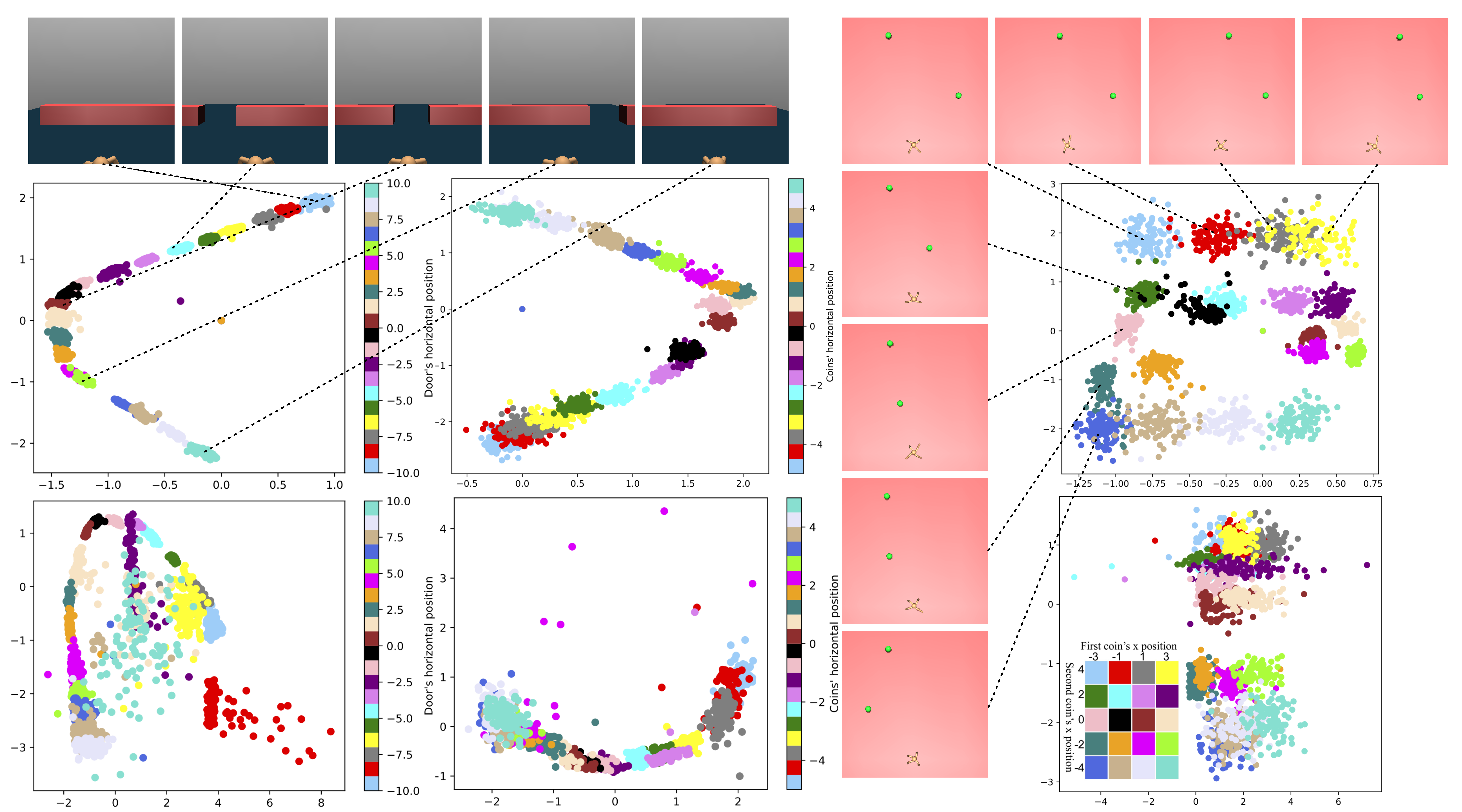}

    \caption{Visualization of learned skill embedding (best among three random seeds) of MLPS (first row) and OPML (second row), from left to right: Ant-Goal, Ant-Gather-one-coin (only one coin's position is changing), Ant-Gather-two-coins. We draw the ground-truth distribution of how the tasks are generated for ant-gather-two-coins at the bottom right corner of the last figure.} 
    \label{fig:visz}
\end{figure*}
\subsection{Overall Performance Comparison}

As shown in Figure~\ref{fig:res1}, we compared to OPML+HPS, which means that we run OPML without the smoothness term to learn the parameterized skills and use our proposed higher-level algorithm HPS to learn the policy. As mentioned before, we use PEARL as the OPML baseline, and we further augment it with contrastive loss as suggested by~\citep{DBLP:conf/aaai/FuTHCFLL21}. We also compared to MLPS+HyAR and MLPS+PDQN, which means that we use the same parameterized skills learned by MLPS but use different PAMDP learning algorithms to do high level policy learning.

As shown in the attached video, the agent trained by our MLPS+HPS algorithm is able to successfully complete the long-horizon tasks in both cases. From Figure~\ref{fig:res1} second row, we can see that the performance drops if we replace the parameterized skills learned by MLPS with that of OFML. The performance gap is much larger than each single skill's performance gap as we will show later (Figure~\ref{fig:metares}), indicating that the proposed trajectory-centric smoothness learning objective help construct a better parameterized action space (Figure~\ref{fig:visz}) which leads to better performance of high-level control policy. With the same pretrained parameterized skills, HPS learns the high-level control policy more efficiently than the other two PAMDP algorithms. In Ant obstacle course tasks, PDQN reaches similar performance in the end but took twice as many environment steps compared to HPS due to the redundancy problem we explained in Section~\ref{sec32}. HyAR fails to learn a good policy possibly because our parameterized action space is learned and synthesized so the noise of high level dynamics is magnified when planning in the further generated latent action space.

\subsection{Quality of the Learned Skill Parameters Space}
We show the visualization of the learned skill parameters' embeddings in Figure~\ref{fig:visz}. For each domain, We run the learned policies on 40 test tasks multiple times to collect enough successful trajectories covering the whole hidden-parameter space. The test tasks are linearly sampled from the given task distribution. Then we encode the trajectories into latent embeddings using the trained context encoder. The original dimension of the latent skill is set to be 2 in the ant domains so we just directly plot the latent embeddings in a 2-D space. As shown in Figure~\ref{fig:visz}, the embeddings generated from the trajectories of the same tasks are close together in the latent space. Moreover, we can see a strong monotonic relationship between the components of the learned latent representation and the real position of the open space in Ant-goal, as well as the coin's horizontal position in Ant-gather. A similar conclusion can also be made in the Ant-gather-two-coins domain, where there are actually \textbf{two} variables for different tasks unlike the other three tasks, which only have one. We can see that the two dimensions of the latent skill approximate these two variables separately, showing a linear correlation between each coin's position and the value of the latent representation. We also compared it with a visualization of the latent embedding encoded using PEARL's context encoder. Without the proposed trajectory-centric smoothness objective, the learned skill embeddings have large areas of overlap and ignore important patterns in the trajectories influenced by the changing positions of the goals.  
\subsection{Quality of the Learned Skill-conditioned Policies}
\begin{figure}[htbp]
\centering
    \includegraphics[width=0.96\linewidth]{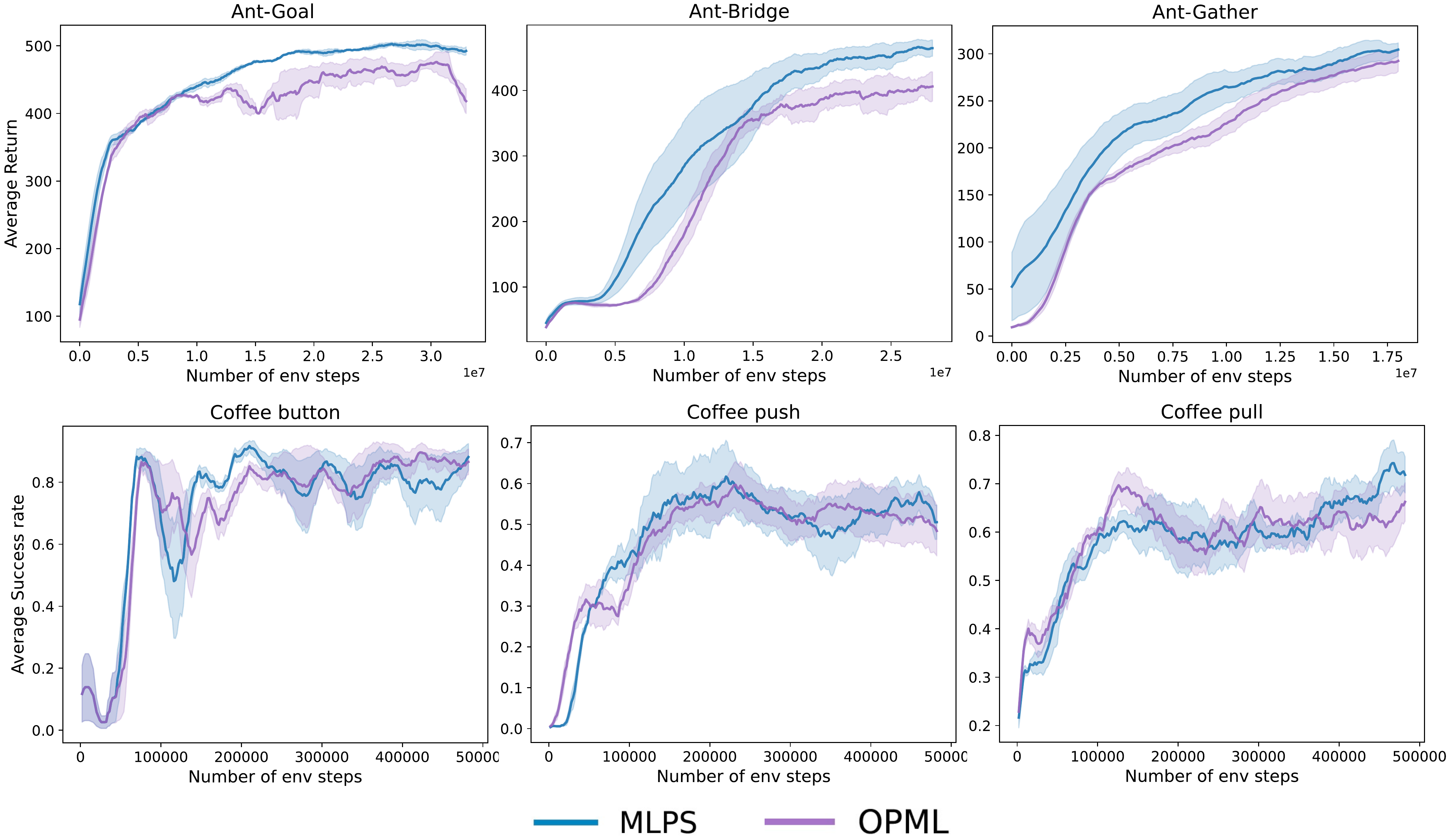}
    \caption{The meta-learning performance comparison on tasks in Ant and Coffee domains.} 
    \label{fig:metares}
\end{figure}
We also compare the performance of MLPS and OPML using standard meta-test in meta-RL to see how the proposed trajectory-centric smoothness objectives in MLPS will influence the low-level skill-conditioned policies' performance. For meta-testing, the test tasks are sampled from the same distribution as the training tasks. The results of meta-testing performance are shown in Figure~\ref{fig:metares}. We find that the smoothness loss does not make the meta-policy's performance worse in any of the tested domains, and actually helps improve the meta-RL performance in the tasks in Ant domain. Unlike the benchmark mujoco tasks in previous meta-RL papers, the difference between optimal policies in these Ant tasks are mainly from the trajectories as a whole, instead of the terminal states/goals. In such settings, which are also common in practice, our proposed trajectory-centric smoothness objectives can help the agent encode the  difference in trajectories into the latent embeddings, thus enabling the agent to quickly identify the correct embedding when adapting to a new task.
\begin{figure*}[htbp]
\centering
    \includegraphics[width=0.96\linewidth]{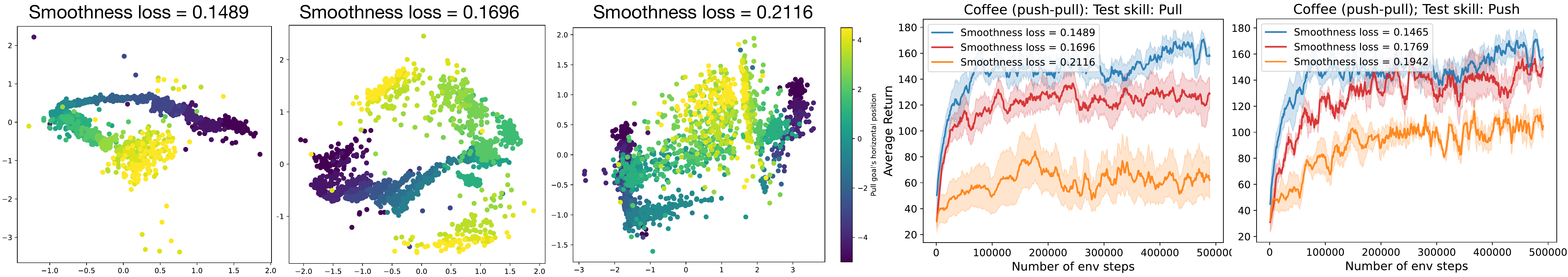}

    \caption{Smoothness of the learned action space and its influence. The left three figures are visualizations of the learned skill embeddings for the \emph{pull} skill with different smoothness loss. Among them, the first two are generated by MLPS and the last one is generated by OPML. The right two figures show the overall performance comparisons of how the smoothness of one skill (pull, push) will affect the agent's overall performance on the long-horizon task. We keep the other skill policy fixed while testing each one of them.} 
    \label{fig:smooth}
\end{figure*}
\subsection{The Importance of Smoothness of the Action Space}
\label{exp:smooth}
In this subsection, we show how the smoothness of the learned action space (skill parameter space) will affect the performance of the policy that will use this action space. We create another coffee long-horizon task where the agent needs to constantly to push and pull the mug to different locations, such that the quality of any one of the two skills will affect the agent's final performance greatly as it has to calculate the skill parameter multiple times within one episode. We use MLPS and OPML to generate a set of \emph{coffee-push} policies and \emph{coffee-pull} policies. We choose those policies with close meta-test success rate to do the further comparison. Then we calculate the normalized smoothness loss for each of them following Equation~(\ref{eqn:dtw}). And we run HPS for each of them and compare their overall performance. We first visualize the influence of smoothness loss to the skill parameter embeddings. For push and pull skill, we set the dimension of the latent parameters as $4$, so we first run Multidimensional Scaling (MDS) and then draw the scatter plot. As shown in Figure~\ref{fig:smooth}, the embedding with the lowest smoothness loss shows the strongest correlation with respect to the change of the real pull target position with few outliers (Color changes from yellow to blue means the target position changes from $-4$ to $4$). And as the smoothness loss increases, more datapoints are dispersed and the correlation becomes weaker. As shown in the right two plots, bad smoothness can greatly increase the difficulty of finding the optimal policy. We find that for both skills we test, the performance of the overall algorithm {\bf drops} fast as the smoothness loss of the learned skill embeddings {\bf increases}, which indicates that smoothness is a very important factor to consider if we are trying to synthesize a new action space composed of skills learned on the primitive action space.

\section{Related Work}

Learning skills in a multi-task setting is common in prior work~\citep{Heess2016LearningAT, Riedmiller2018LearningBP, Hausman2018LearningAE}. \citet{DBLP:conf/icml/SilvaKB12} first proposed to construct parameterized skills by analyzing the structure of policy manifold, but required labeled parameters of tasks for training.
With the meta-RL setting, MLSH~\citep{Frans2018MetaLS} learns fixed low-level policies during training and further finetune the high level policy on new tasks. \citet{Nam2022SkillbasedML} focus on using skilled pretrained from offline data to do better meta-RL. \citet{DBLP:conf/nips/HarrisonSFP20} propose an interesting way to do online changepoint detection in continual learning. Our method, in comparison, assumes we know exactly when the new task arrives during the test phase and we focus on the reinforcement learning setting. Some approaches also introduce multiple levels of hierarchies for skill learning~\citep{CoReyes2018SelfConsistentTA} or planning~\citep{Nachum2018DataEfficientHR}. \citet{Barreto2019TheOK} and \citet{Qureshi2020ComposingTP} propose a method to compose new task-relevant skills with pretrained simple skills. \citet{Goyal2020ReinforcementLW} learn a high-level controller with decentralized low-level policies. However, these low-level skills are not parameterized so the generalization ability is limited. \citet{Rao2021LearningTM} introduce a similar three-level hierarchy of policies that also have discrete and continuous parts. However, they focus on learning skills from offline dataset and the learned skills do not involve temporal abstraction of the actions.  Another category of skill learning method is unsupervised skill discovery~\citep{DBLP:conf/iclr/Kwon21, DBLP:conf/iclr/EysenbachGIL19,Campos2020ExploreDA, DBLP:conf/icml/BagariaS021}. In particular, DADS~\citep{DBLP:conf/iclr/SharmaGLKH20} successfully encode trajectories into a smooth latent skill pace in simple navigation tasks. 
However, the pure unsupervised learning setting does not allow the agent to master one complete category of high-level skill, e.g., find the coffee machine in a house, because of lack of task-specific exploration as no environmental reward is given. Thus it’s hard to directly use these skills to solve long-horizon tasks.

A large body of recent work focuses on Deep RL problems with parameterized action spaces~\citep{DBLP:conf/aaai/MassonRK16}. We have discussed PDQN~\citep{Bester2019MultiPassQF, Xiong2018ParametrizedDQ} and HyAR~\citep{Li2021HyARAD} in previous sections. PADDPG~\citep{DBLP:journals/corr/HausknechtS15a} and HPPO~\citep{Fan2019HybridAR} let the actor output an concatenation of the discrete action and the continuous parameters for each discrete action label together. This category of methods tends to ignore the dependency between discrete action and continuous parameter, which is crucial for finding the optimal parameterized action. 
\citet{Neunert2019ContinuousDiscreteRL} also considers discrete-continuous control but the settings are not standard parameterized action space, i.e., the discrete part and the continuous part of action are independent of each other. 
Parameterized actions have also been studied in the task and motion planning (TAMP) literature~\citep{DBLP:conf/icra/KaelblingL11, DBLP:conf/nips/DalalPS21, DBLP:conf/icra/ChitnisT0020, Silver2022PredicateIF, DBLP:conf/iros/SilverCTKL21, DBLP:conf/icra/NasirianyLZ22}. These approaches typically assume the parameterzed skills already exist.
By contrast, our three-level hierarchy policies are all learned from scratch using RL.

\section{Conclusion}
\label{concl}
We propose a three-level hierarchy framework that models a temporally-extended PAMDP. 
We leverage off-policy Meta-RL framework to learn the skills while further augment it with a trajectory-centric smoothness loss to train the trajectory encoder -- aiming to improve the smoothness of the latent parameter space. We empirically show that our meta-learning parameterized skills framework enables an agent to solve two sets of complex long-horizon continuous control tasks. We also demonstrate the importance of the different components of our algorithm independently.

\section*{Acknowledgement}
The authors would like to thank Akhil Bagaria, Sam Lobel, Anita de Mello Koch, Paul Zhiyuan Zhou and other members of Brown bigAI, as well as Tom Silver, Rohan Chitnis, Riley Simmons-Edler, Anurag Ajay for discussions and helpful feedback, and the anonymous reviewers for valuable feedback that improved the paper substantially. This work was supported in part by an NSF Graduate Research Fellowship under grant \#2040433, NSF grants \#1717569 \#1955361 and CAREER award \#1844960, DARPA grant W911NF1820268, ONR contracts N00014-17-1-2699 and N00014-22-1-2592, and the DARPA Lifelong Learning Machines
program under grant \#FA8750-18-2-0117. This work was conducted using computational resources and services at the Center for Computation and Visualization, Brown University.

\bibliography{reff}
\bibliographystyle{icml2023}

\newpage
\appendix
\onecolumn
\section{Appendix}

\subsection{Meta-Learning Parameterized Skill (MLPS) Algorithm}
\label{app:algo}

\begin{algorithm}[htbp]
\begin{algorithmic}

\STATE {\bfseries Input:} Batch of training tasks ${\mu}_{i=1,\cdots, M}$ from $p(\mu)$, 
\STATE Initialize replay buffer $B_i$ for each training task
\STATE Initialize parameters $\theta_{a}$ and $\theta_{c}$ for the actor and critic networks separately.
\STATE Initialize parameters context encoder network $\phi$, context encoder target network $\phi_{target}$
\WHILE{not done}
\FOR{each task $\mu_i$}
\STATE Roll out policy $\pi_{\theta_a}$, producing transitions $\{(s_j, a_j, r_j, s'_j)\}_{j:1\cdots N}$
\STATE Add tuples to execution replay buffer $B_i$
\ENDFOR
\IF{there's at least one success trajectory in each task's replay buffer}
\STATE $calculating\textunderscore DTW=True$
\ENDIF
\FOR{each training step}
\STATE Sample a meta batch of tasks $\{1, \cdots, C\}$
    \FOR{each task $i$ in meta batch}
    \STATE Sample two transition batches $b^i_1 = \{(s_k, a_k, r_k, s'_k)\}_{k = 1\cdots K} \sim B_i$, $b^i_2 = \{(s_k, a_k, r_k, s'_k)\}_{k = 1\cdots K} \sim B_i$ 
    
    \STATE Sample latent embedding $z^i_1 \sim \phi(b^i_1)$, $z^i_{target} \sim \phi_{target}(b^i_2)$
    
    \STATE Update actor and critic networks with $\{z^i_1, b^i_1\}$, and calculate $L_{Value}$
    \ENDFOR
    \STATE Calculate contrastive loss $L_{NCE}$ with $\{z^1_1, \cdots, z^C_1\}$, $\{z^1_{target}, \cdots, z^C_{target}\}$
    \IF{$calculating\textunderscore DTW=True$}
    \STATE Sample one success trajectory from each task's replay buffer: $\{\tau_{suc}^1, \cdots, \tau_{suc}^C\}$
    \STATE Calculate Dynamic Time Warping loss $L_{Smoothness}$ with $\{z^1_1, \cdots, z^C_1\}$, $\{z^1_{target}, \cdots, z^C_{target}\}$, $\{\tau_{suc}^1, \cdots, \tau_{suc}^C\}$  
    \ENDIF
    \STATE Update cotext encoder network with $L_{Skill} = L_{Value} + \alpha L_{NCE} + \beta L_{Smoothness}$
    
    \ENDFOR

\ENDWHILE

 \end{algorithmic}
 \caption{Meta-Learning Parameterized Skill (MLPS) Meta-training (regular encoder network)} \label{pslalgo}
\end{algorithm}

We show detailed procedures in Algorithm~\ref{pslalgo}. The training procedures for the actor and critic networks are the same as in PEARL. After collecting data, for each training step, we first sample a meta batch of tasks $\{1, \cdots, C\}$. Then for each task, we sample two transition batches $b^i_1$ and $b^i_2$ from its own replay buffer. We feed the first transition batch into the context encoder, then use the output latent embedding to calculate the RL loss $L_{Value}$ and update actor and critic network parameters. This procedure is the same as in PEARL. We feed the second transition batch into the target context encoder network to get the latent embedding which will be used to calculate the auxiliary losses. After we get all the latent embeddings for tasks in the meta batch, we first calculate the contrastive loss using the latent embedding pairs from given task set. Then, if each task has collected at least one success trajectory (that is, the agent successfully reached the goal position), we will let the agent also calculate Dynamic Time Warping loss with the latent embedding and success trajectories sampled for each task in the meta batch. And we will update the context encoder network's parameters at the end of this training step. Note that one limitation of the implementation here is that for some tasks, it is possible that not all tasks in the training task set can collect a success trajectory within the given number of episodes. This will lead to the problem that the DTW is not calculated and used throughout the training process. Thus, we provide another implementation in~\ref{option}, which does not have such requirement and achieves similar final performance.

For calculating contrastive loss, we adopt the same procedures in~\citep{DBLP:conf/icml/LaskinSA20,DBLP:conf/aaai/FuTHCFLL21}, where we model the similarity score calculating function as bilinear products, i.e. $z_{\mu}^TWz_{k}$, where $W$ is the learned parameter. Using the denotations in Algorithm~\ref{pslalgo}, for $z^1_1$, we can rewrite the InfoNCE loss as:
\[
    L_{NCE} := - \mathbb{E}[f(z^1_{1}, z^1_{target}) - \log \frac{1}{N}\sum_{j=2}^C \exp(f(z^1_{1}, z^j_{target})))].
\]
And we calculate the loss use same procedures for other latent embedding $\{z^2_1,\cdots, z^C_1\}$. 

For calculating Dynamic Time Warping loss, given a latent embedding pair from different tasks: $(z^j_1, z^k_{target})$, we draw the corresponding pair from the success trajectories set: $(\tau^j_{suc}, \tau^k_{suc})$, and calculate the DTW loss with:
\begin{equation}
    L_{\Smoothness} := \mathbb{E}_{\tau^j_{suc}, \tau^k_{suc}}MSE[||z^j_1 - z^k_{target}||_{2}- \kappa \DTW(\tau^j_{suc}, \tau^k_{suc})],
\label{equa}
\end{equation}
where $\kappa$ denotes the hyperparameter controls the scale of the DTW distance.

Different from standard meta-RL setting, we assume the training task set (a fixed number of tasks) is given, whereas in~\citep{Finn2017ModelAgnosticMF} each time a task is randomly generated using parameters sampled from a prior distribution.
\begin{table}[htb]
\caption{MLPS's hyperparameters}
\centering
\begin{tabular}{lllllll}
\centering

 Environment & \# Meta-train tasks & $\alpha$ & $\beta$ &$\kappa$ & Meta batch size & Embedding batch size \\\hline 
 Ant-goal & 100 & 10 & 1 & 0.5 & 16 & 100\\
 Ant-bridge & 100 & 100 & 1 & 0.1 & 16 & 50\\
 Ant-gather-one-coin & 100 & 10 & 1 &0.5 & 16 & 100\\
 Ant-gather-two-coins & 200 & 10 & 0.1 & 0.5&32 & 150\\
 Coffee-push & 60 & 0 & 0.1 &0.5 & 16 & 100\\ 
 Coffee-pull & 60 & 0 & 0.1 &0.5 & 16 & 100\\
 Coffee-button & 60 & 0 & 0.1 &0.5 & 16 & 100\\\hline

\end{tabular}

\label{tab1}
\end{table}

\subsubsection{Implementation details}
When computing the latent embedding $z$ using context encoder, for coffee domain, the state component in the input trajectory only contains the first three elements (x\&y\&z coordinates of the gripper). For ant domain, the state component in the input trajectory only contains the first two elements (x\&y coordinates of the ant) for ant-goal and ant-gather, for ant-bridge the state component in the input trajectory is the original state. Both actor network and critic network in MLPS are parameterized MLPs with 2 hidden layers of $(300, 300)$ units. The context/trajectory encoder network is modeled as product of independent Gaussian factors, with 3 hidden layers of $(400, 400, 400)$ units. We set the learning rate as $3e-4$. The scale of KL divergence loss is set to be $0.1$. Other hyperparameters are listed in Table~\ref{tab1}.

\subsubsection{Another approach for implementing the context encoder and its training process}
\label{option}
\begin{wrapfigure}{r}{0.36\textwidth}
    \includegraphics[width=1\linewidth]{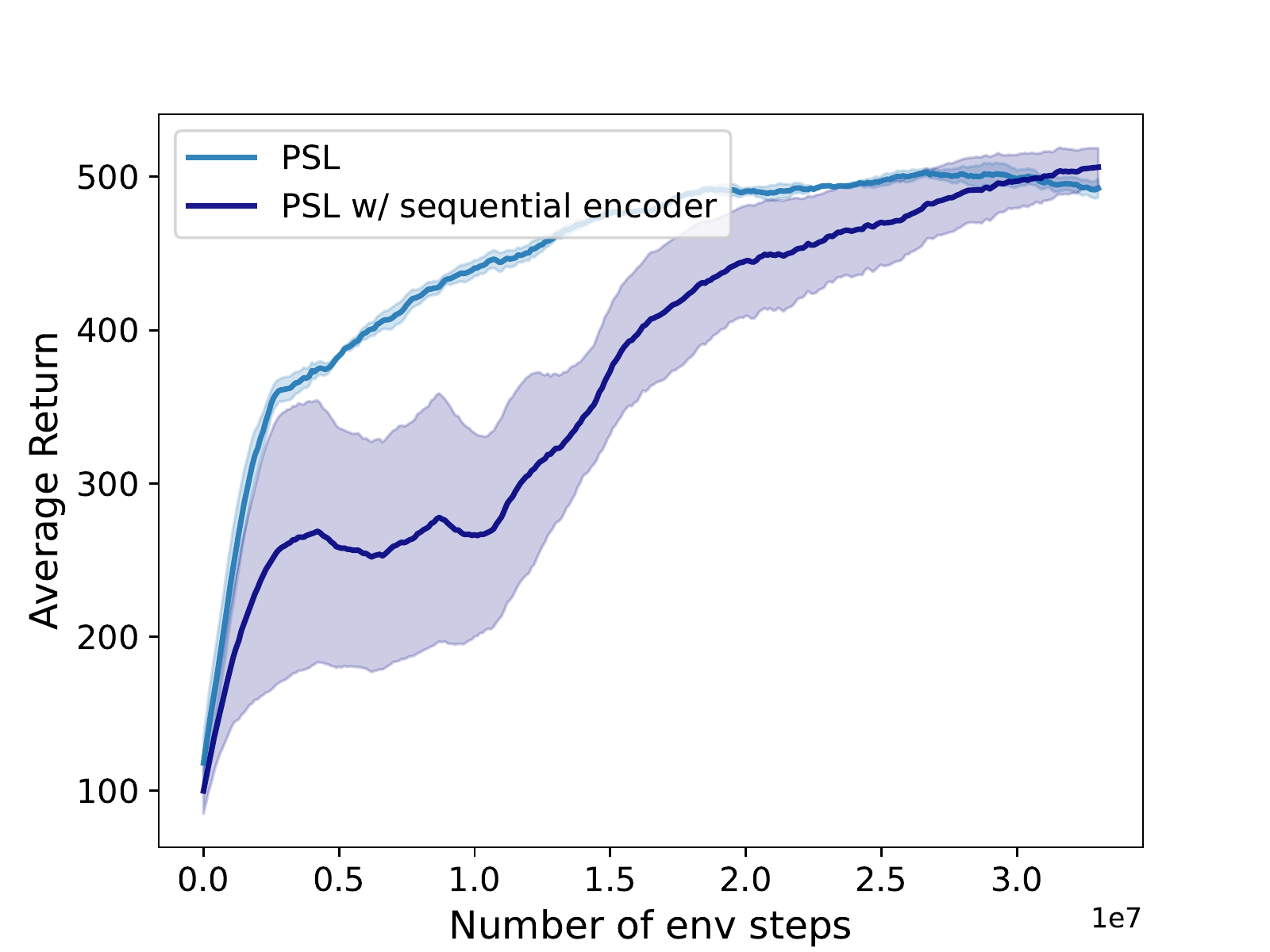}
    \caption{Comparison of different implementation strategy for MLPS on Ant-goal.} 
    \label{fig:att}
\end{wrapfigure}
Based on the intuition that the distance of different skills in the latent space should be proportional to the distance between their
trajectories, we can compute DTW distance for any pair of trajectories, no matter if they succeed or not, and match the distance to their corresponding latent embeddings' distance. Thus, we do not need to wait until there's at least one success trajectory in each task's replay buffer to calculate the smoothness loss.

Concretely, we provide the algorithm in Algorithm~\ref{pslalgo2}. Instead of modeling the context/trajectory encoder network as a product of independent Gaussian factors, we use a sequential encoder network, SNAIL~\citep{DBLP:conf/iclr/MishraR0A18}, which uses temporal convolution and soft attention. Then, at each training step, instead of sampling two random batches of {\bf transitions}, we sample two complete {\bf trajectories} $\tau_1, \tau_2$ and transform them to the same length. We compute the corresponding latent embeddings $z_1, z_2$ for both of them using the context encoder, and calculate the DTW distance as well as the smoothness loss using the same equation~(\ref{equa}). Thus we update the encoder network with the smoothness loss at every training step. 

We show the results comparison in Figure~\ref{fig:att}. Although MLPS with sequential encoder does not learn as fast as the original version, it achieves similar final performance. Besides, the requirement for using this version of the algorithm is a little looser. The readers can choose to apply one of the two versions of our algorithm based on the properties of their own test tasks.

\begin{algorithm}[htbp]
\begin{algorithmic}

\STATE {\bfseries Input:} Batch of training tasks ${\mu}_{i=1,\cdots, M}$ from $p(\mu)$, 
\STATE Initialize replay buffer $B_i$ for each training task
\STATE Initialize parameters $\theta_{a}$ and $\theta_{c}$ for the actor and critic networks separately.
\STATE Initialize parameters context encoder network $\phi$, context encoder target network $\phi_{target}$
\WHILE{not done}
\FOR{each task $\mu_i$}
\STATE Roll out policy $\pi_{\theta_a}$, producing transitions $\{(s_j, a_j, r_j, s'_j)\}_{j:1\cdots N}$
\STATE Add tuples to execution replay buffer $B_i$
\ENDFOR
\IF{there's at least one success trajectory in each task's replay buffer}
\STATE $calculating\textunderscore DTW=True$
\ENDIF
\FOR{each training step}
\STATE Sample a meta batch of tasks $\{1, \cdots, C\}$
    \FOR{each task $i$ in meta batch}
    \STATE Sample two trajectories and transform them to same length $K$:  $\tau^i_1 = \{(s_k, a_k, r_k, s'_k)\}_{k = 1\cdots K}\sim B_i$, $\tau^i_2 = \{(s_k, a_k, r_k, s'_k)\}_{k = 1\cdots K}\sim B_i$ 
    
    \STATE Sample latent embedding $z^i_1 \sim \phi(\tau^i_1)$, $z^i_{target} \sim \phi_{target}(\tau^i_2)$
    \STATE Sample transition batch $b^i = \{(s_k, a_k, r_k, s'_k)\}_{k = 1\cdots K} \sim B_i$
    \STATE Update actor and critic networks with $\{z^i_1, b^i\}$, and calculate $L_{Value}$
    \ENDFOR
    \STATE Calculate contrastive loss $L_{NCE}$ with $\{z^1_1, \cdots, z^C_1\}$, $\{z^1_{target}, \cdots, z^C_{target}\}$
    
    \STATE Calculate Dynamic Time Warping distance and Smoothness loss $L_{Smoothness}$ with $\{z^1_1, \cdots, z^C_1\}$, $\{z^1_{target}, \cdots, z^C_{target}\}$, $\{\tau_{1}^1, \cdots, \tau_{1}^C\}$, $\{\tau_{2}^1, \cdots, \tau_{2}^C\}$   

    \STATE Update context encoder network with $L_{Skill} = L_{Value} + \alpha L_{NCE} + \beta L_{Smoothness}$
    
    \ENDFOR

\ENDWHILE

 \end{algorithmic}
 \caption{Parameterized Skill Learning (MLPS) Meta-training (Sequential encoder network)} \label{pslalgo2}
\end{algorithm}


         

\subsection{Hierarchical actor-critic with Parameterized Skills (HPS)}
\subsubsection{Further Comparison with other existing RL with parameterized action space algorithms}
We show a comparison of different algorithms' properties in Table~\ref{tab2}. P-DQN lacks scalability as it maintains a separate actor network for each discrete action, and have to compute all of them during both training and execution as we explained in the main text. HHQN has the problem of potential nonstationarity as we explained in the last paragraph of Section 4.2. PADDPG makes the actor output an concatenation of the discrete action and the continuous parameters for each of them together, which tends to ignore the dependency between discrete action and continuous parameters. This leads to performance drop as shown in PDQN and HyAR's original papers. HyAR don't have the above three problems but it needs to further learn a latent action space and plan based on it instead of the primitive parameterized action space. In our scenario where the parameterized action space is actually learned, the noise in the dynamics is magnified and it's hard to learn a proper latent action space. We assume this leads to HyAR's performance drop in our experiments.

\begin{table}[htb]
\centering
\caption{Comparison with other parameterized action space algorithms}
\begin{tabular}{lllll}
\centering

 Algorithm & Scalability & Stationarity & Dependence & Primitive  \\\hline 
 P-DQN & \XSolidBrush & \checkmark & \checkmark & \checkmark \\
 PADDPG & \checkmark & \checkmark & \XSolidBrush & \checkmark \\
 HHQN & \checkmark & \XSolidBrush & \checkmark & \checkmark \\
 HyAR & \checkmark & \checkmark & \checkmark & \XSolidBrush \\\hline
  HPS & \checkmark & \checkmark & \checkmark & \checkmark \\\hline

\end{tabular}

\label{tab2}
\end{table}
\subsubsection{Implementation details}
For the actor of discrete action $\pi_{\theta_d}$, we use two hidden layers of MLPs with $(300, 300)$ units, the output layer follows by a gumbel-softmax layer. For both the actor of continuous parameters $\pi_{\theta_c}$ and critic network, we use two hidden layers of MLPs with $(300, 300)$ units. The learning rates are all set as $3e-4$. The output of the actor of continuous parameters are stochastic the same as in SAC. Note that we fix the temperature for gumbel-softmax to be $1.0$ across the whole training process, without using any decaying strategy. We also tried automatic temperature tuning as in SAC but did not get satisfactory result. We set the reward scale as $5$ and the batch size as $128$.

\subsubsection{Temporally-extended PAMDP}
\begin{wrapfigure}{r}{0.46\textwidth}
    \includegraphics[width=1\linewidth]{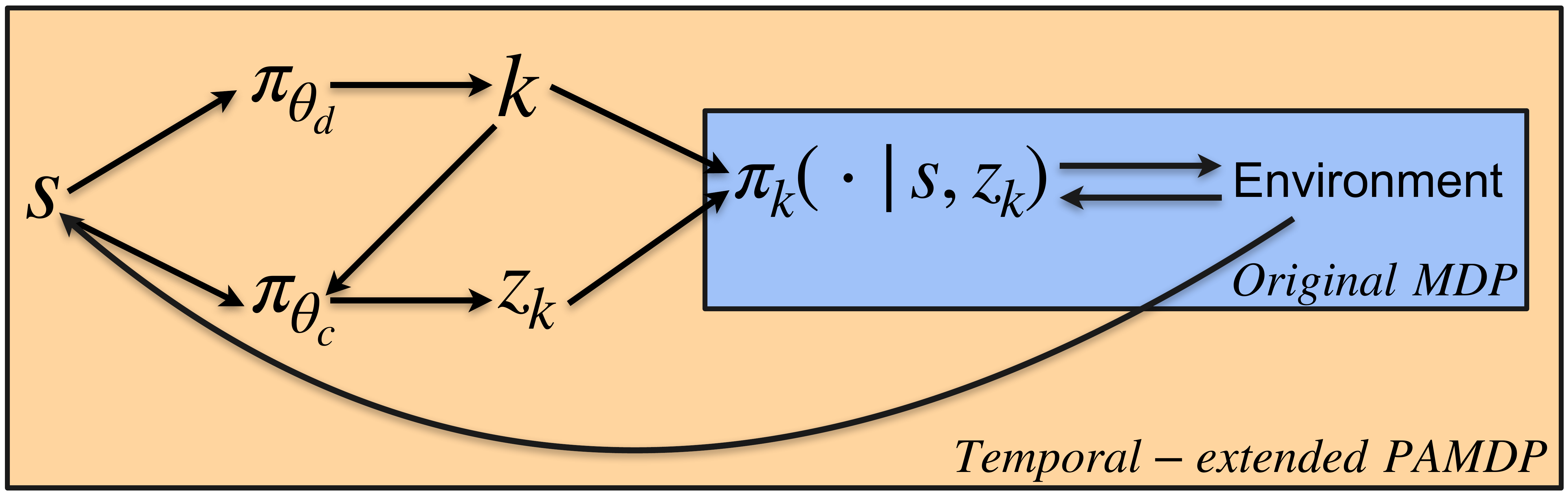}
    \caption{Decision making process in the temporally-extended PAMDP. $\pi_{\theta_d}$ denotes the policy for discrete action and $\pi_{\theta_c}$ denotes the policy for continuous parameters.} 
    \label{fig:psp}
\end{wrapfigure}
Specifically, after we let the agent train on $K$ different categories of tasks using MLPS, we get $K$ different skill-conditioned policies and fix them. Then we can directly let the high-level agent solve a new task by learning a policy that maps states to parameterized skill pairs $(k, z_k)$---learning in the high-level temporal-extended action space. Each discrete skill label corresponds to a low-level skill-conditioned policy network $\pi_k(a|s, z_k)$, which takes the continuous skill parameter $z_k$ as an additional input. The decision making process of this new temporal-extended PAMDP is illustrated in Figure~\ref{fig:psp}. Upon receiving a new observation, the agent must first choose the discrete skill label $k$ using $\pi_{\theta_d}$ and then choose the corresponding skill parameter $z_k$ given the state $s$ and $k$ using $\pi_{\theta_c}$. The low-level skill-conditioned policy $\pi_k(a|s, z_k)$, which is learned by MLPS and fixed, takes in the observation and the skill parameter and outputs a primitive action to interact with the environment. The discrete skill label and the continuous parameter are fixed and the low level policy $\pi_k(a|s, z_k)$ will constantly output actions for a given number of environmental steps. Then, the last observation received from the environment is used as the new input state for the high-level policy, which will select new $k$ and $z_k$, and so on. 

\subsection{Environment details and baselines}
\label{app:env}
We run all experiment with the mujoco simulator~\citep{DBLP:conf/iros/TodorovET12}:
\begin{itemize}
    \item Ant-goal: The horizontal position of the doorway changes across all the tasks (uniformly sampled from $[-10,10]$). The other environmental properties are fixed, including the goal's position. The task horizon is 400. The agent succeeds when it reaches the goal position $(x=0, y=25)$. The state input includes the position and velocity of different joints of ant, and the ant's horizontal position $x$, as well as its relative vertical position $y$ to the midlane $y=10$.
    
    \item Ant-bridge: The wind speed when the ant is on the bridge changes across all the tasks (uniformly sampled from $[-3,3]$). The other environmental properties are fixed. The task horizon is 300. The agent succeeds when it reaches the goal position $(x=0, y=26)$. The state input includes the position and velocity of different joints of ant, and the ant's horizontal position $x$, as well as its relative vertical position $y$ to the midlane $y=10$.
    
    \item Ant-gather: The position of the first coin (Ant-gather-one-coin) or both coins (Ant-gather-two-coins) change across all the tasks (uniformly sampled from $[-4.5,4.5]$). The other environmental properties are fixed\footnote{Note that in this paper, we consider the HiP-MDP setting. If we change the task order as well as the task parameters, without giving the agent these information the problem would become partially observable and extremely hard to solve.}. The task horizon is 400. The agent succeeds only when it gathers both coins and reaches the goal position $(x=0, y=16)$. The state input includes the position and velocity of different joints of ant,  an indicator for how many coins the ant has gathered, and the ant's horizontal position $x$, as well as its relative vertical position $y$ to the midlane $y=8$.
    
    \item Ant-box: The ant needs to push the box and walk pass a gap to reach the goal position. The position of the box is fixed. The task horizon is 500.
    
    \item Ant-mix: The ant needs pass 10/15 different barriers consist of Ant-goal, Ant-bridge, Ant-gather-one-coin, Ant-box and reach the goal position. The task order as well as their specific features (door position, wind speed etc.) are all fixed. The origianl task horizon is 4000/6000. The task horizon when we do high-level learning with the skills is 10/15. The state input includes: High-level: the ant's horizontal position $x$ and vertical position $y$, how many barrier it has passed. Low-level: the ant's horizontal position $x$ and its relative vertical position $y$ to the midlane of the current subtask, as well as the position and velocity of different joints of ant, and how many coins the ant has gathered.

    \item Coffee-button: We adopt the same environment in MetaWorld\citep{DBLP:conf/corl/YuQHJHFL19}. The goal is press a button on the coffee machine. The button's position if changing across different tasks.

    \item Coffee-push: We adopt the same environment in MetaWorld and further modify it by letting the gripper start at a position above the mug at the beginning of every episode. The goal is to push the mug to a target position under the coffee machine. The target position is changing across different tasks.

    \item Coffee-pull: We adopt the same environment in MetaWorld and further modify it by letting the gripper start at a position above the mug at the beginning of every episode. The goal is to pull the mug to a target position under the coffee machine. The target position is changing across different tasks.

    \item Reach: The goal is to reach the mug. This is a discrete skill.

    \item Return: The goal is to return to the gripper's start position. This is a discrete skill. 
\end{itemize}
Reward Functions:
\begin{itemize}
    \item Ant-goal:
    \begin{equation*}
    \begin{aligned}
    R_{t} = &\mathbb{I}\{\text{The ant has not passed the door}\} * \Delta d_{\text{Distance to door}} + \mathbb{I}\{\text{door}\} * 10\\ &+ \mathbb{I}\{\text{The ant has passed the door}\}* \Delta d_{\text{Distance to goal}} + \mathbb{I}\{\text{goal}\} * 20
    \end{aligned}
    \end{equation*}
    \item Ant-bridge:
    \begin{equation*}
    \begin{aligned}
    R_{t} = \Delta d_{\text{Distance to goal}} + \mathbb{I}\{\text{goal}\} * 20
    \end{aligned}
\end{equation*}
\item Ant-gather:
    \begin{equation*}
    \begin{aligned}
    R_{t} = &\mathbb{I}\{\text{The ant has not gathered the first coin}\} * \Delta d_{\text{Distance to first coin}} + \mathbb{I}\{\text{first coin}\} * 10\\ &+ \mathbb{I}\{\text{The ant has gathered one coin, one left}\}* \Delta d_{\text{Distance to second coin}} + \mathbb{I}\{\text{second coin}\} * 10 \\ & \mathbb{I}\{\text{The ant has gathered two coins}\}* \Delta d_{\text{Distance to goal}} + \mathbb{I}\{\text{goal}\} * 20
    \end{aligned}
    \end{equation*}
\item Ant-mix (sparse):
    \begin{equation*}
    \begin{aligned}
    R_{t} = \mathbb{I}\{\text{The ant passed a barrier}\} * 5+ \mathbb{I}\{\text{goal}\} * 100
    \end{aligned}
    \end{equation*}
    
\item Ant-mix (dense): For the dense reward used by other baselines, we use the direct combination of the dense reward we set for each specific subtask. The environment knows what the subtask is and it will give the corresponding dense reward. Moreover, we also give it the sparse reward when it passes each barrier.

\item Coffee-button, coffee-push, coffee-pull: same as in the original MetaWorld.
\end{itemize}

The number of environment steps needed to complete the tasks (Ant-mix) and reach the final goal is around 3500 for 10b-3c, and around 5000 for 15b-4c.

The results shown in the main text are averaged over three random seeds. The error bar shows one standard deviation. All experiments were run on our university's high performance computing cluster. When comparing with PDQN \& HyAR \& PEARL in ant obstacle course (ant-mix) domain (results shown in two plots of Figure 8), we fix the task order across different random seeds to make the environment setting consistent to all baselines. 

\begin{figure}[htbp]
\centering
    \includegraphics[width=0.7\linewidth]{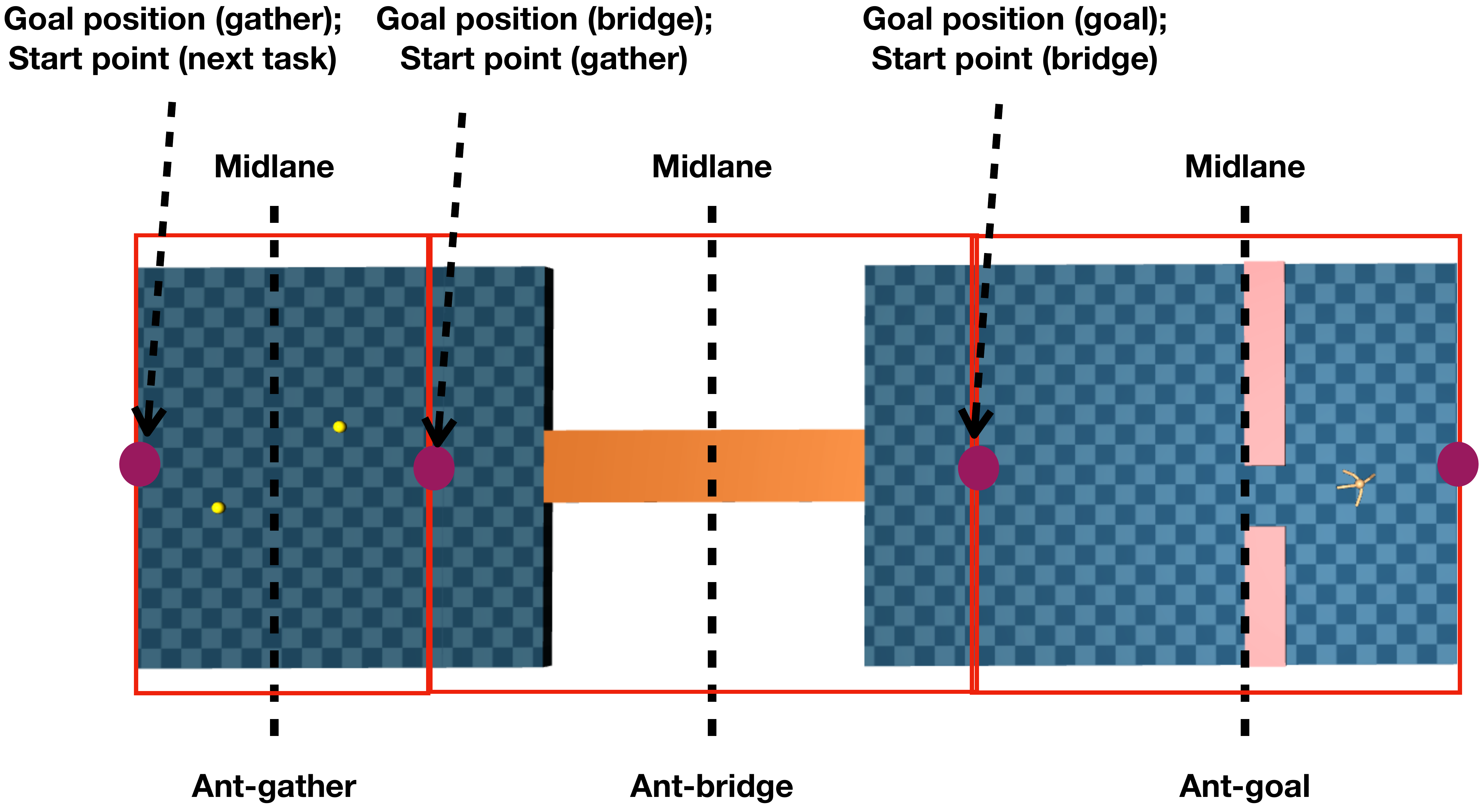}
    \caption{Ant obstacle course (ant-mix) further illustration.} 
    \label{fig:goalpos}
\end{figure}


Baselines: 1. Parameterized skill learning: We use the original source code for PEARL\footnote{\url{https://github.com/katerakelly/oyster}}, VariBAD\footnote{\url{https://github.com/lmzintgraf/varibad}} and their implementation for $\text{RL}^2$. 2. Learning with learned parameterized skills: We use the original code for HyAR-TD3\footnote{\url{https://github.com/TJU-DRL-LAB/AI-Optimizer/tree/1e2a33a4a3a7a8235f1c12ea71b1ea686c071094/self-supervised-rl/RL_with_Action_Representation/HyAR}}, and their implementation for PDQN-TD3. For SAC, we use the stable-baselines3 implementation\footnote{\url{https://github.com/DLR-RM/stable-baselines3}}\citep{stable-baselines3}. Additionally, for HyAR, we let the agent pretrain the Variational Auto-encoder 2000 steps. 

\subsection{More Experimental results}
\label{app:exp}
We compared the difference between DTW distance and pointwise euclidean distance. For each domain (ant-goal, ant-bridge, ant-gather), we test two scenarios: {\bf same tasks}, where we fix the hidden parameter (door position/wind speed/coin position), and calculate the distance between success trajectories that are able to solve the same task. Another scenario is {\bf neighbour tasks}, where we sample $5$ values from the original range of the hidden parameter with same distance from each other. For instance, for ant-goal, we sample $5$ doorway position: $\{-9, -4.5, 0, 4.5, 9\}$. Then we calculate the distance between success trajectories from two neighbour tasks. Ideally, the distance of different pairs of neighbour tasks (e.g. $\{-9, -4.5\} \& \{-4.5, 0\}$) should be similar to each other, as the actual distance between the hidden parameters are the same.

\begin{figure}[htbp]
\centering
    \includegraphics[width=0.15\linewidth]{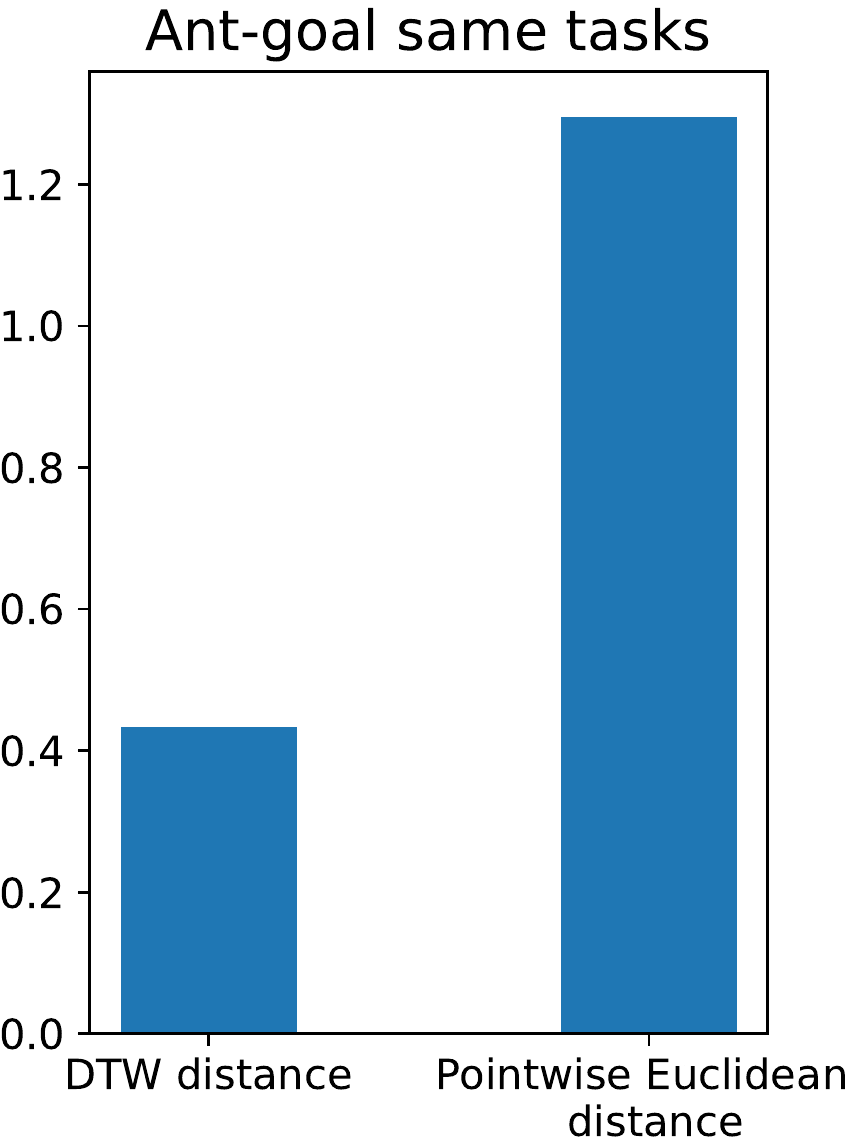}
    \includegraphics[width=0.15\linewidth]{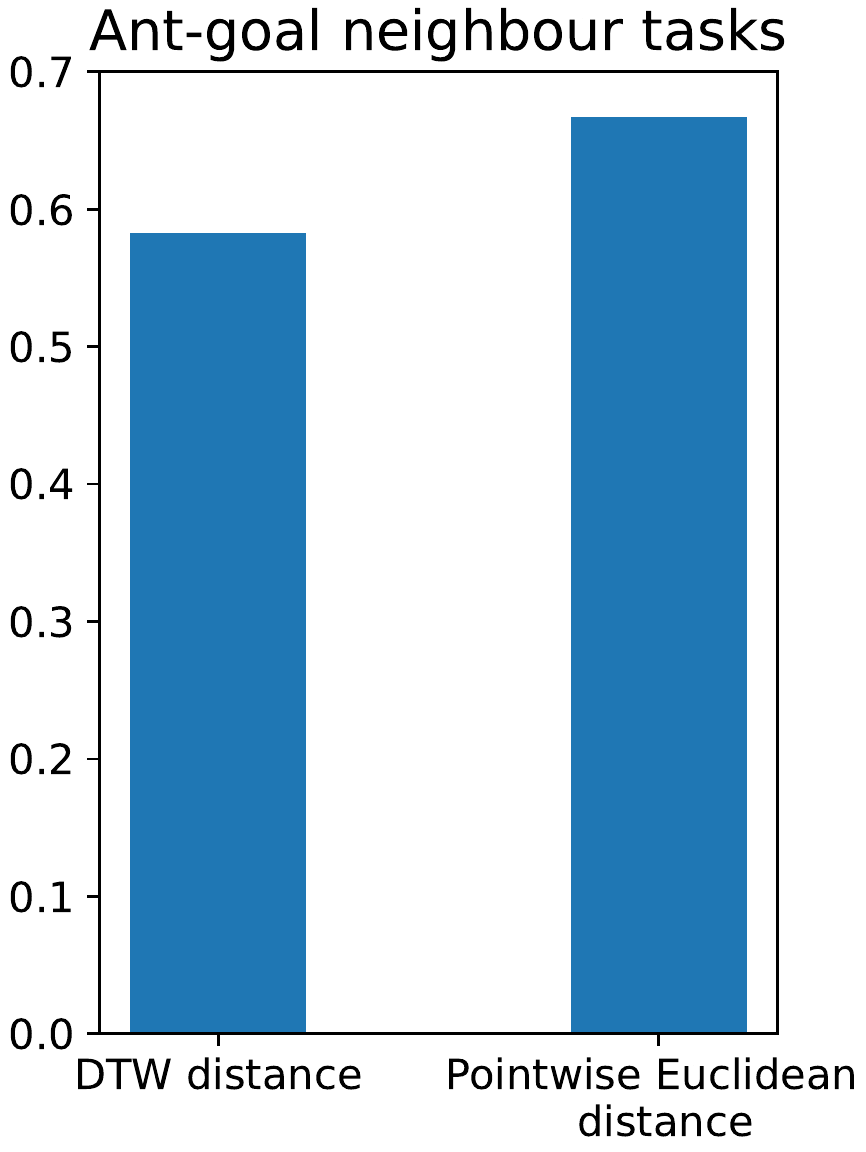}
    \includegraphics[width=0.16\linewidth]{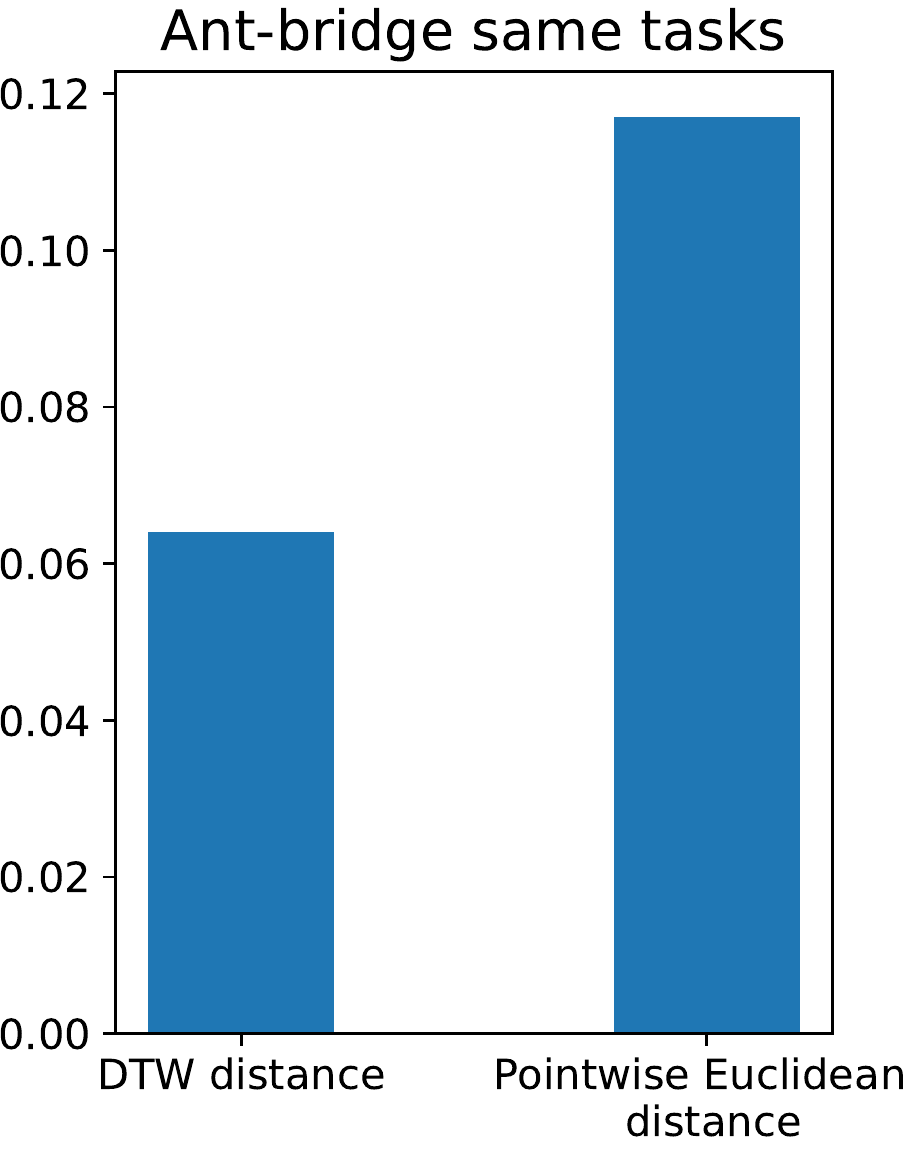}
    \includegraphics[width=0.16\linewidth]{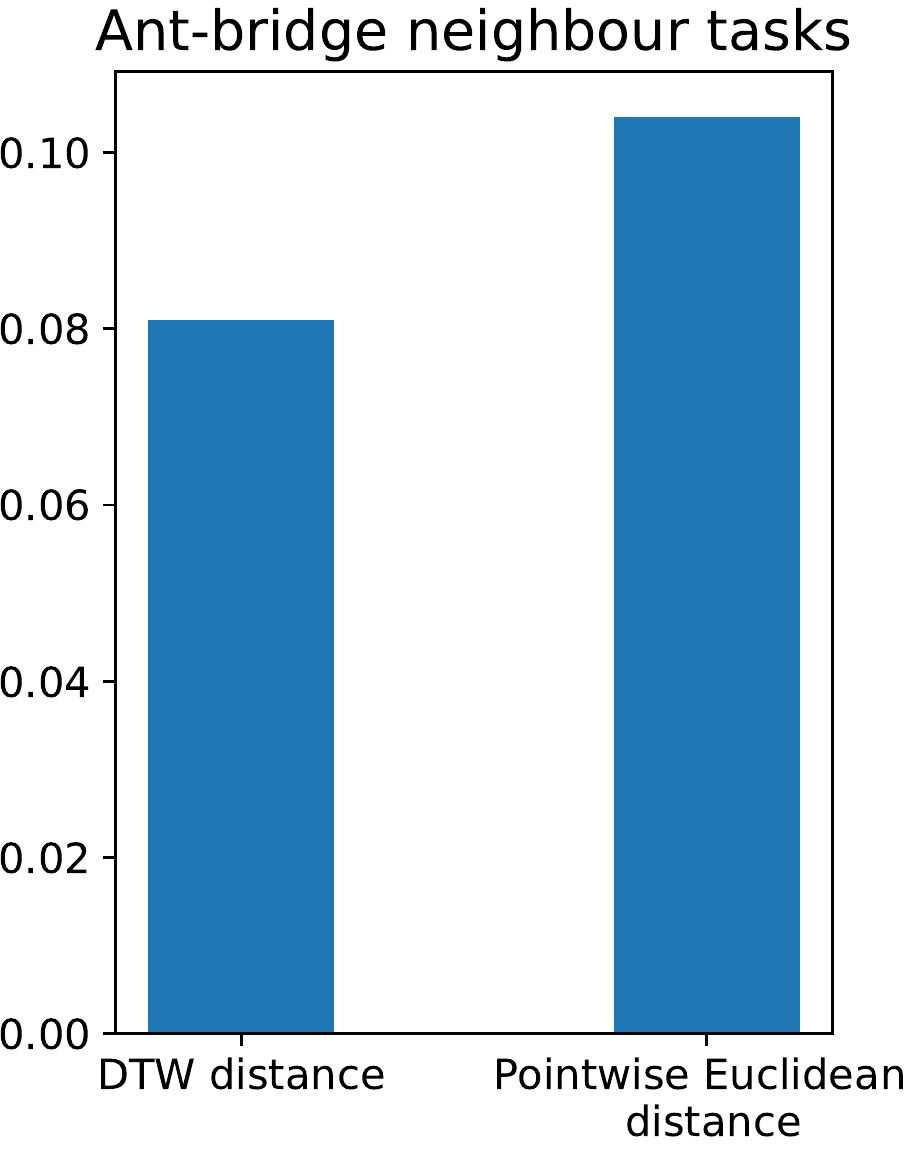}
    \includegraphics[width=0.157\linewidth]{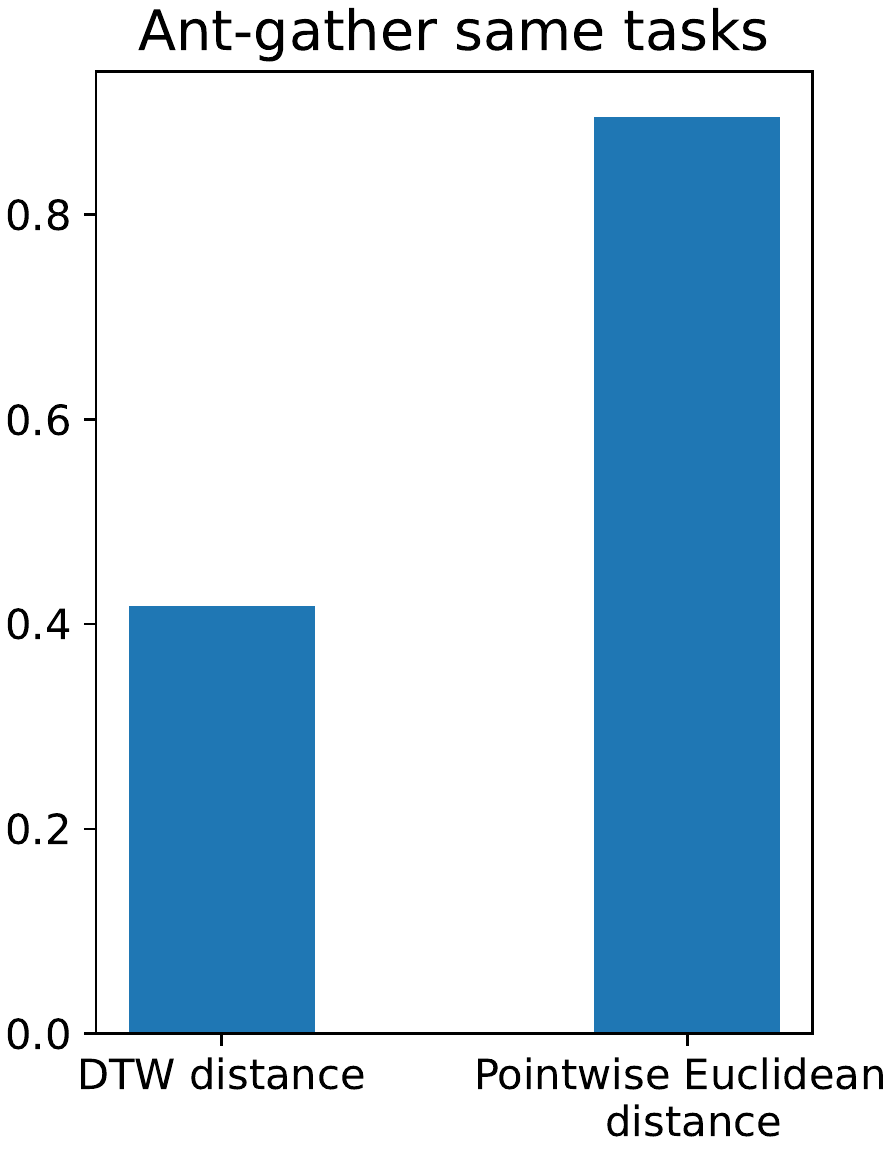}
    \includegraphics[width=0.16\linewidth]{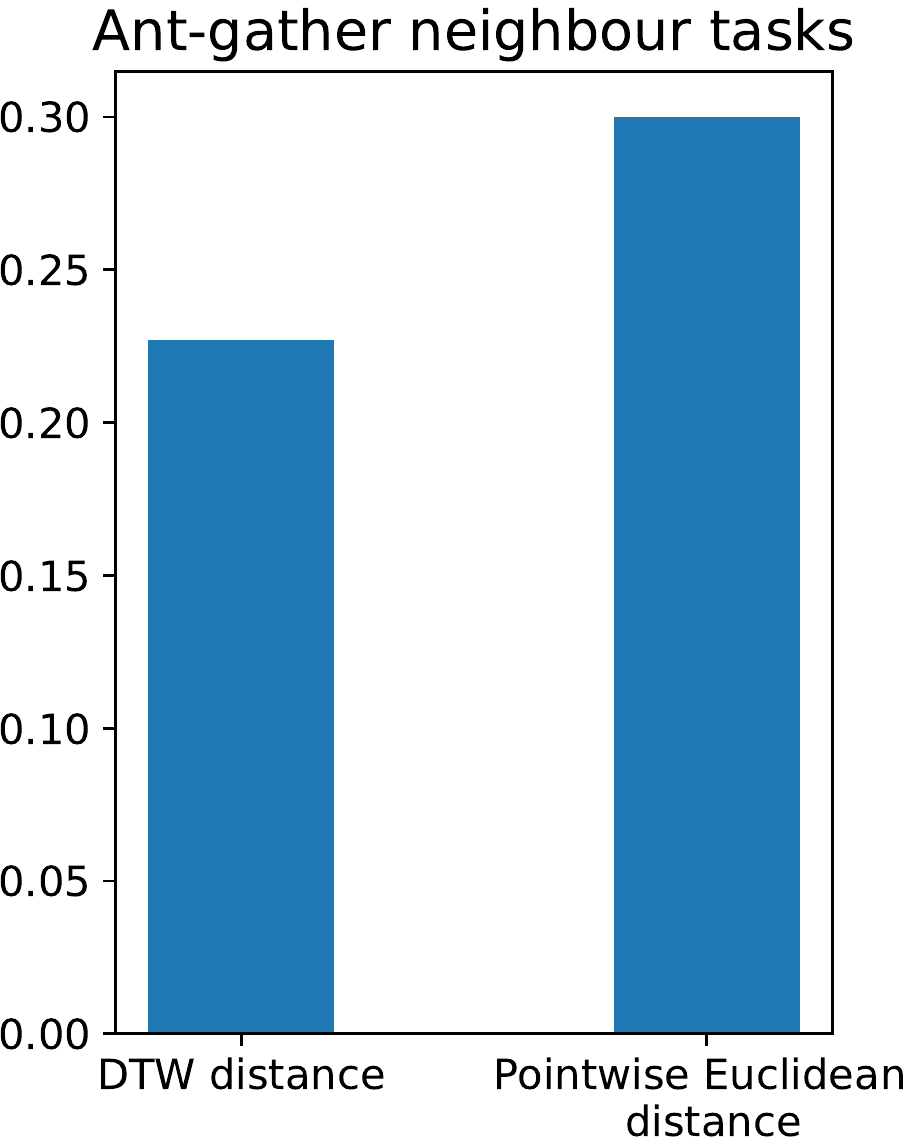}
    \caption{Comparison results of DTW distance against pointwise euclidean distance in {\bf Coefficient of Variation}. In all scenarios, we collect ten pairs of data (two categories of distance) and then compute the coefficient of variation for the ten values.} 
    \label{fig:CV}
\end{figure}

We show the Coefficient of Variation of the two methods for calculating distance in different scenarios in Figure~\ref{fig:CV}. In both same tasks and neighbour tasks scenarios, we expect the coefficient of variation to be small. This is because different metrics will result in different means, but the variation of the distance should be small as these distance are either calculated for the same tasks (that is, actual hidden parameter distance is fixed as $0$) or for tasks with the same actual hidden parameter distance. We find that the distance calculated by DTW gets smaller variation in all scenarios which is consistent to our hypothesis. The gap between the two methods is especially large for trajectories from the same tasks, indicating that unwrapped pointwise Euclidean distance can end up with the erroneous conclusion that the trajectories are very different even though they have quite similar overall shape.

\begin{figure}[htbp]
\centering
    \includegraphics[width=0.31\linewidth]{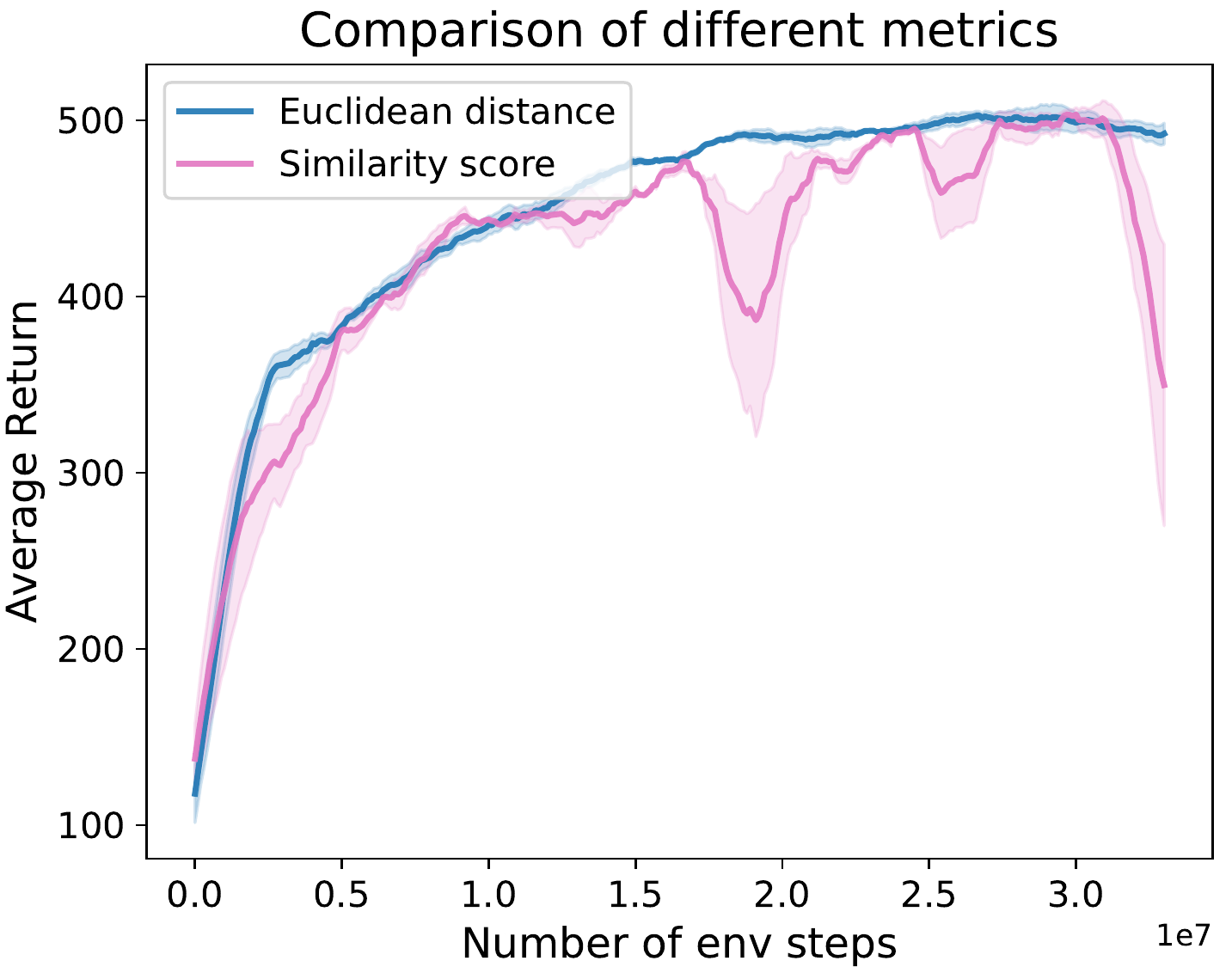}
    \includegraphics[width=0.33\linewidth]{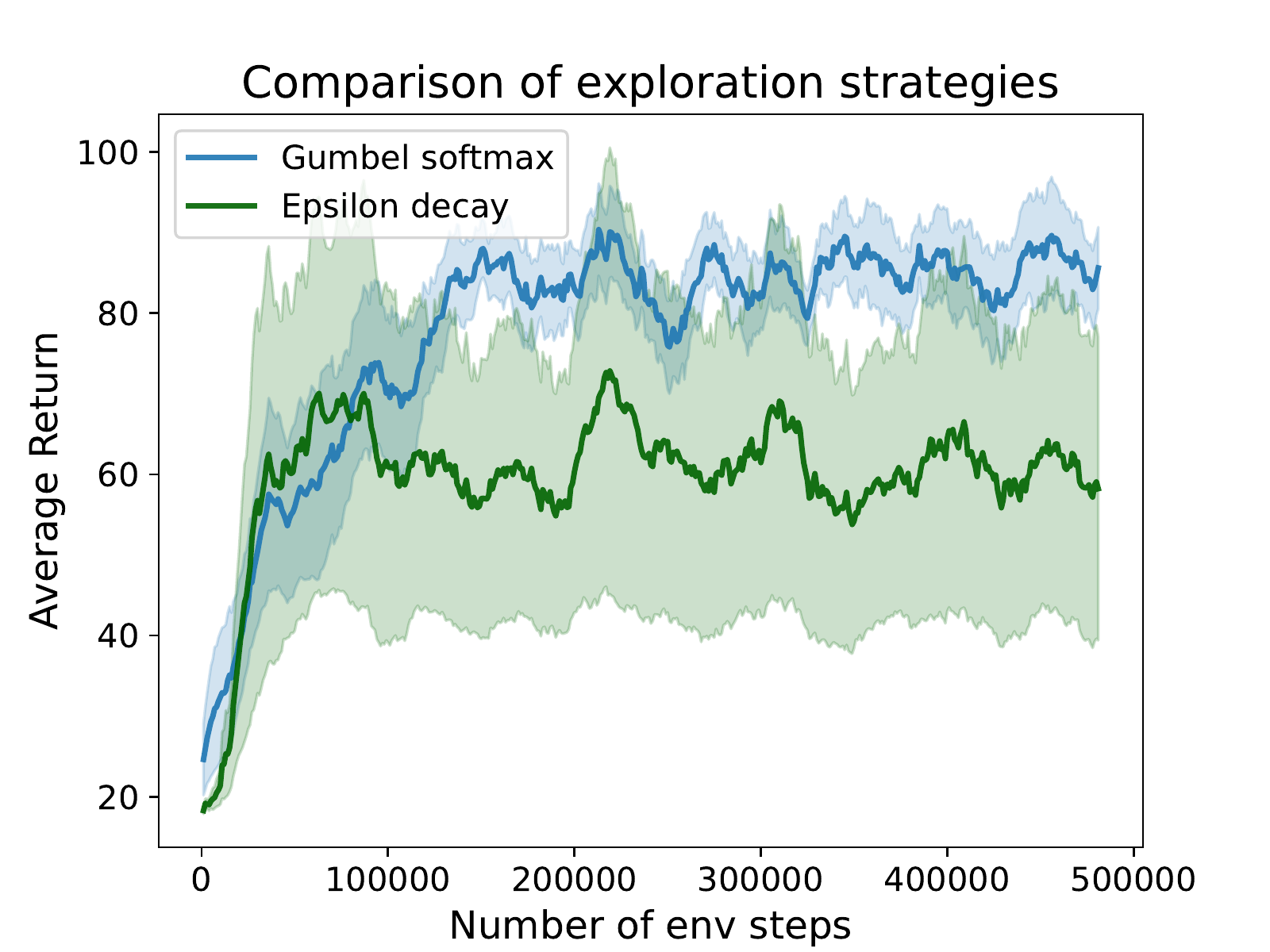}
    \caption{Left: Comparison of different metrics for calculating the distance between latent embeddings on Ant-goal. Right: Comparison of different exploration strategies for \emph{HPS} on Ant obstacle course 10b-3c.} 
    \label{fig:metric}
\end{figure}

We also compare the metrics for calculating the distance between $z$ when calculating the DTW distance, shown in Figure~\ref{fig:metric} Left. We find that besides directly using Euclidean distance as in Equation~(\ref{equa}), we can also use the similarity score function $f$ to calculate the distance between two latent embedding. And the result shows that these two metrics achieve similar results, although the performance of using similarity score drops a bit at the end of training. As shown in Figure~\ref{fig:metric} Right, compared with regular epsilon-greedy strategy, our exploration strategy based on gumbel-softmax is important for HPS to achieve good performance on ant obstacle course tasks. Moreover, we do not need to consider the additional hyperparameters brought by epsilon-greedy method (final epsilon, number of decay steps) and just fix the ``temparture'' of gumbel-softmax to be $1.0$ for all the scenarios. 

\begin{figure}[htbp]
\centering
    \includegraphics[width=0.8\linewidth]{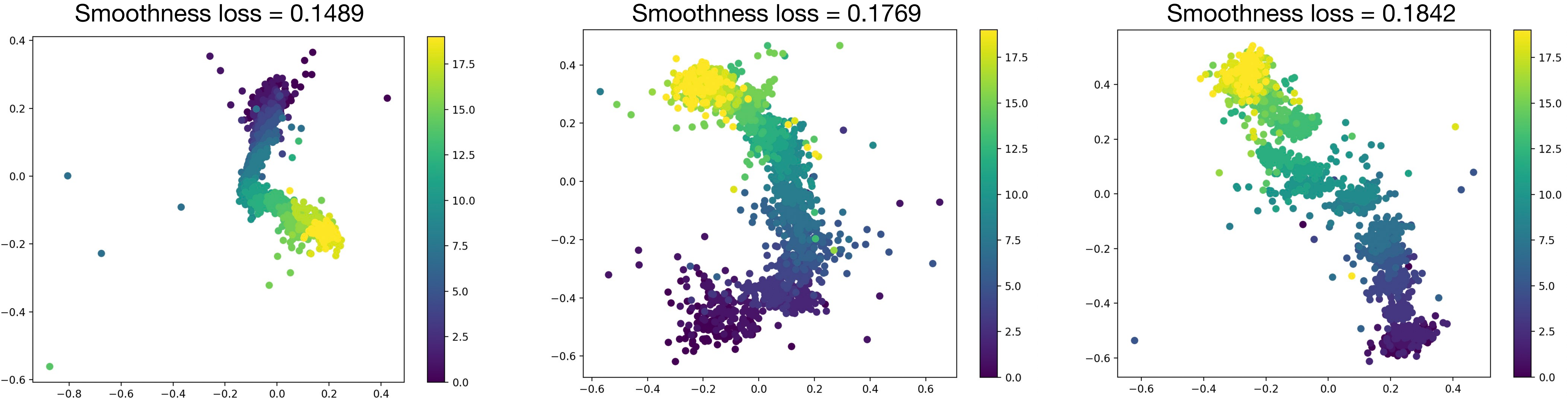}
    \caption{Visualization of coffee-push skill with different smoothness loss (corresponding to Figure~\ref{fig:smooth}).} 
    \label{fig:app1}
\end{figure}

\begin{figure}[htbp]
\centering
    \includegraphics[width=0.8\linewidth]{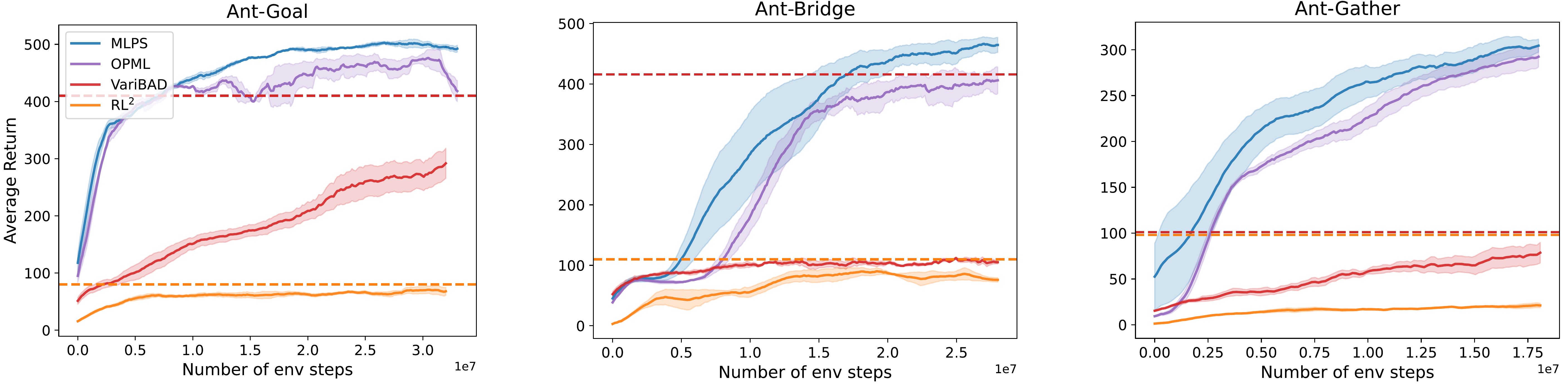}
    \caption{Comparison with On-policy Meta-RL methods on the ant domain.} 
    \label{fig:app2}
\end{figure}
\subsection{Ablation study for HPS and HHQN}
\label{hhqn}
\begin{figure}[htbp]
\centering
    \includegraphics[width=0.3\linewidth]{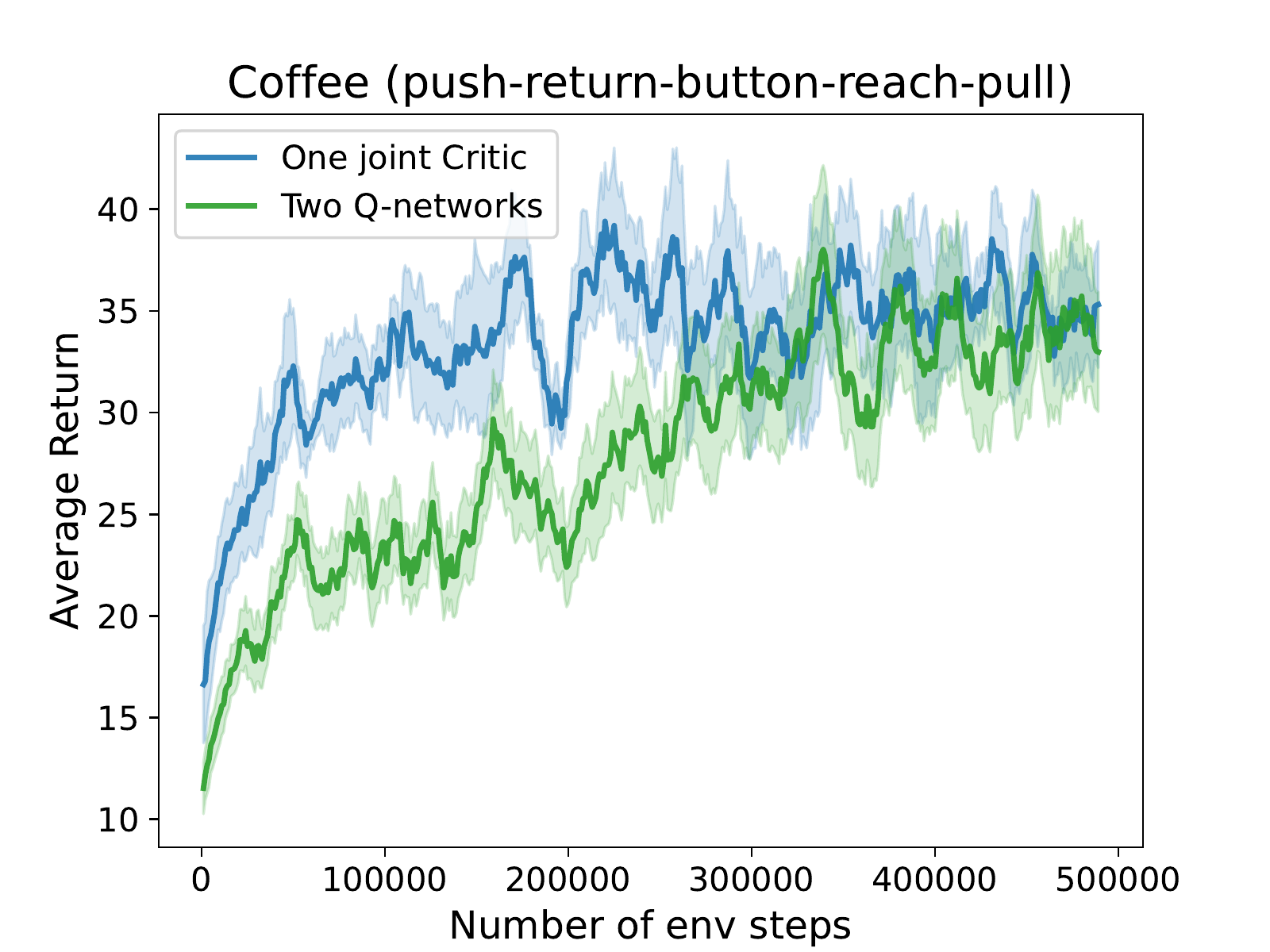}

    \caption{Ablation study for HPS: one joint critic network v.s. two separate critic networks.} 
    \label{fig:hhqn}
\end{figure}

We also conduct an ablation study on using two different Q networks for discrete and continuous skill respectively like HHQN, while keeping the other components the same as our algorithm. As shown in Figure~\ref{fig:hhqn}, in the coffee-make long horizon task, using one joint critic network as in our algorithm HPS results in a faster learning speed than training two Q-networks for discrete and continuous skill parameters separately. For both methods, we use the same set of low-level skills learned by MLPS.

\subsection{Comparison with Hierarchical RL and Skill discovery methods}
\label{hrlbaseline}
\begin{figure}[htbp]
\centering
    \includegraphics[width=0.3\linewidth]{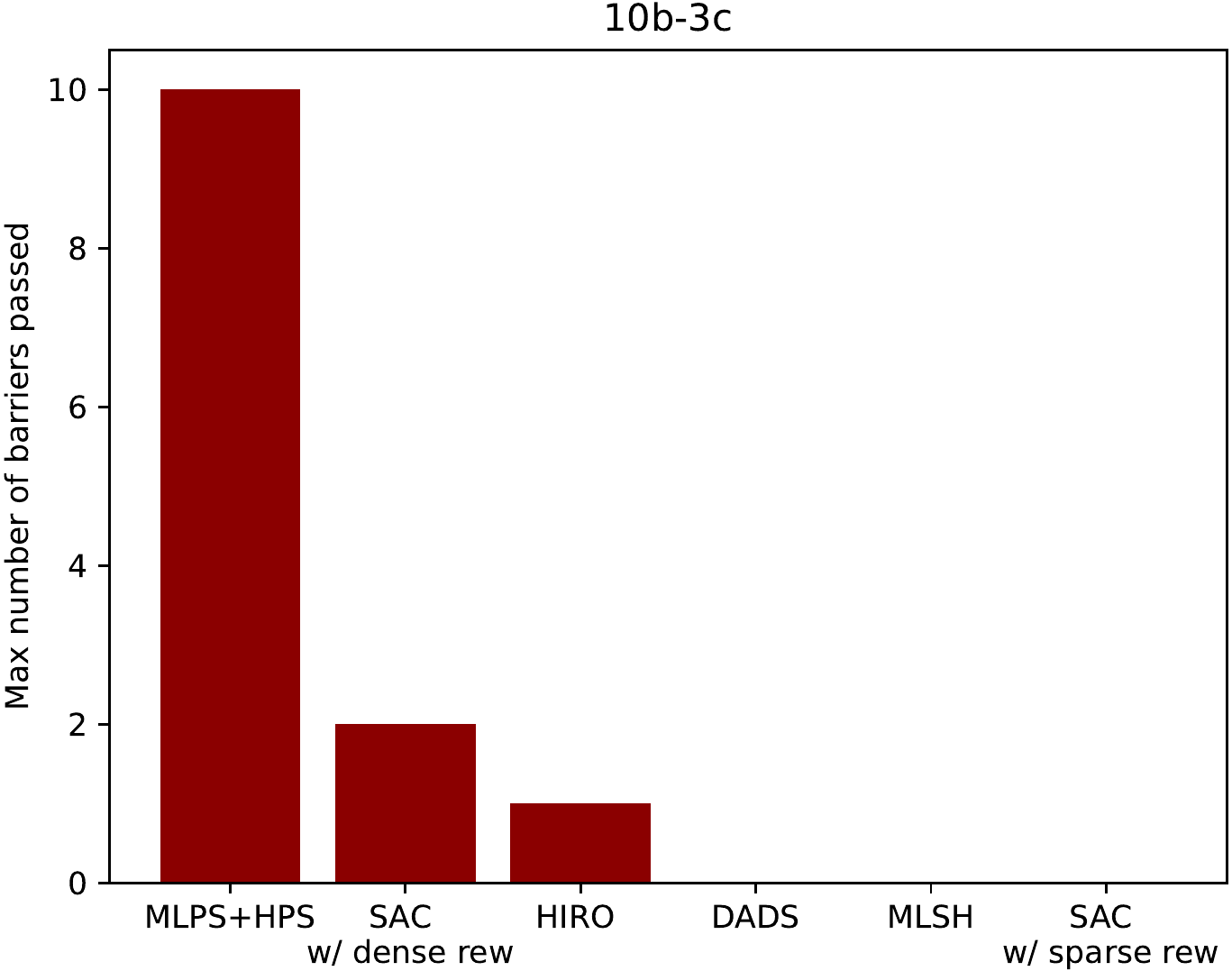}
    \includegraphics[width=0.29\linewidth]{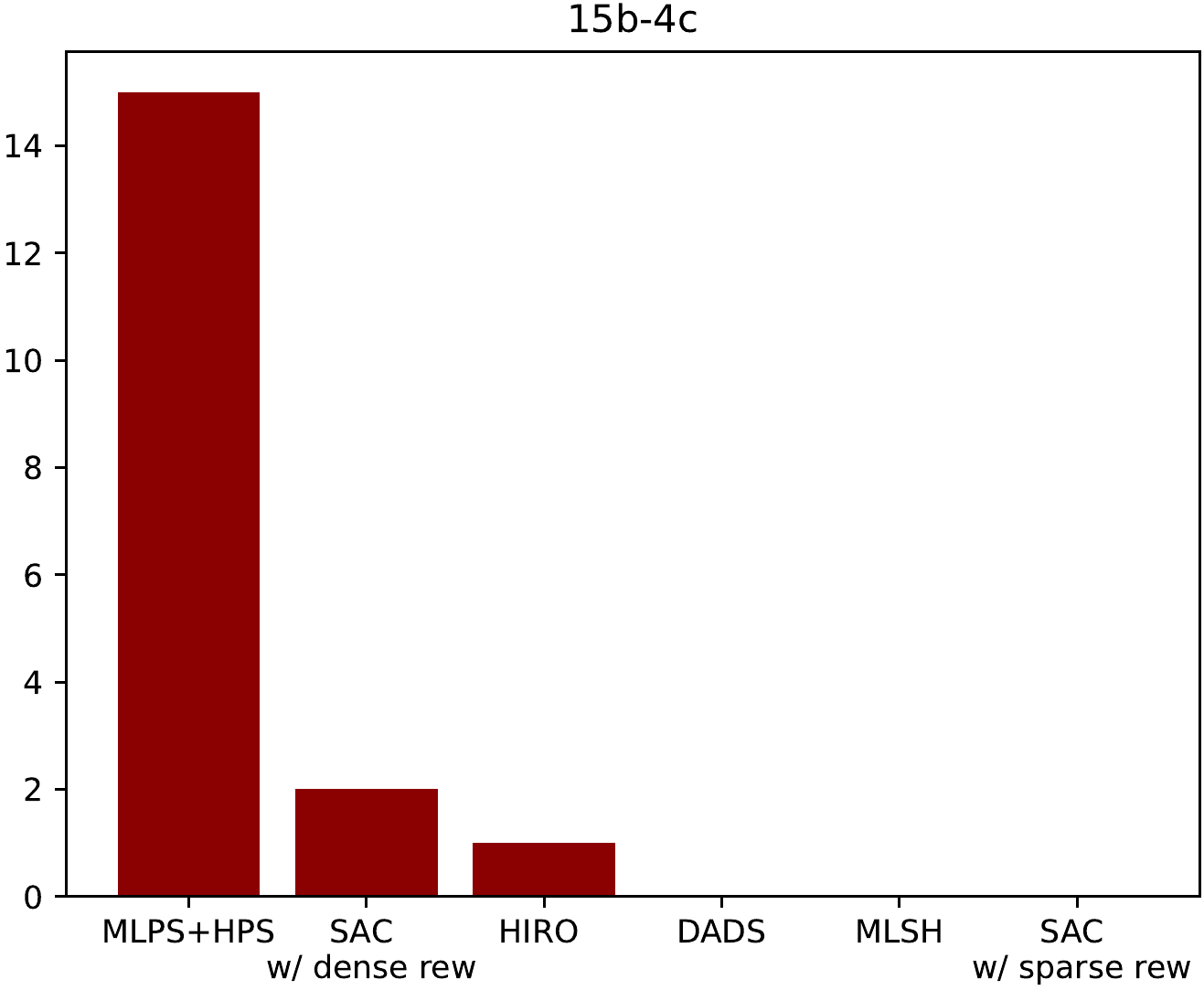}
    \caption{Comparison results of MLPS + HPS against other baselines in ant-mix domain (10b-3c and 15b-4c).} 
    \label{fig:hrl}
\end{figure}

 We run two relevant hierarchical RL methods (HIRO~\citep{Nachum2018DataEfficientHR} and MLSH~\citep{Frans2018MetaLS}) and one skill discovery method (off-policy DADS~\citep{DBLP:conf/iclr/SharmaGLKH20, DBLP:conf/rss/SharmaALKHG20}) on the two ant-mix domain: 10b-3c (10 consecutive barriers sampled from 3 different categories of tasks introduced the previous section) and 15b-3c (15 consecutive barriers sampled from 4 different categories of tasks). We use the official released code for all the baselines\footnote{\url{https://github.com/openai/mlsh}}\footnote{\url{https://github.com/watakandai/hiro_pytorch}}\footnote{\url{https://github.com/google-research/dads}}. We train all these three methods with dense reward like we train SAC from scratch (for our method we used sparse reward). The dense reward is a composition of the exact same dense reward we used to train the low-level parameterized skills. We show the comparison of the results in the ant-mix domain. As shown in Figure~\ref{fig:hrl}, the agent trained by the other algorithms can pass two barriers at most, while our MLPS+HPS algorithm is able to pass all 10/15 barriers. 

\subsection{Experimental results for smoothly changing subtasks}

\begin{figure}[htbp]
\centering
    \includegraphics[width=0.36\linewidth]{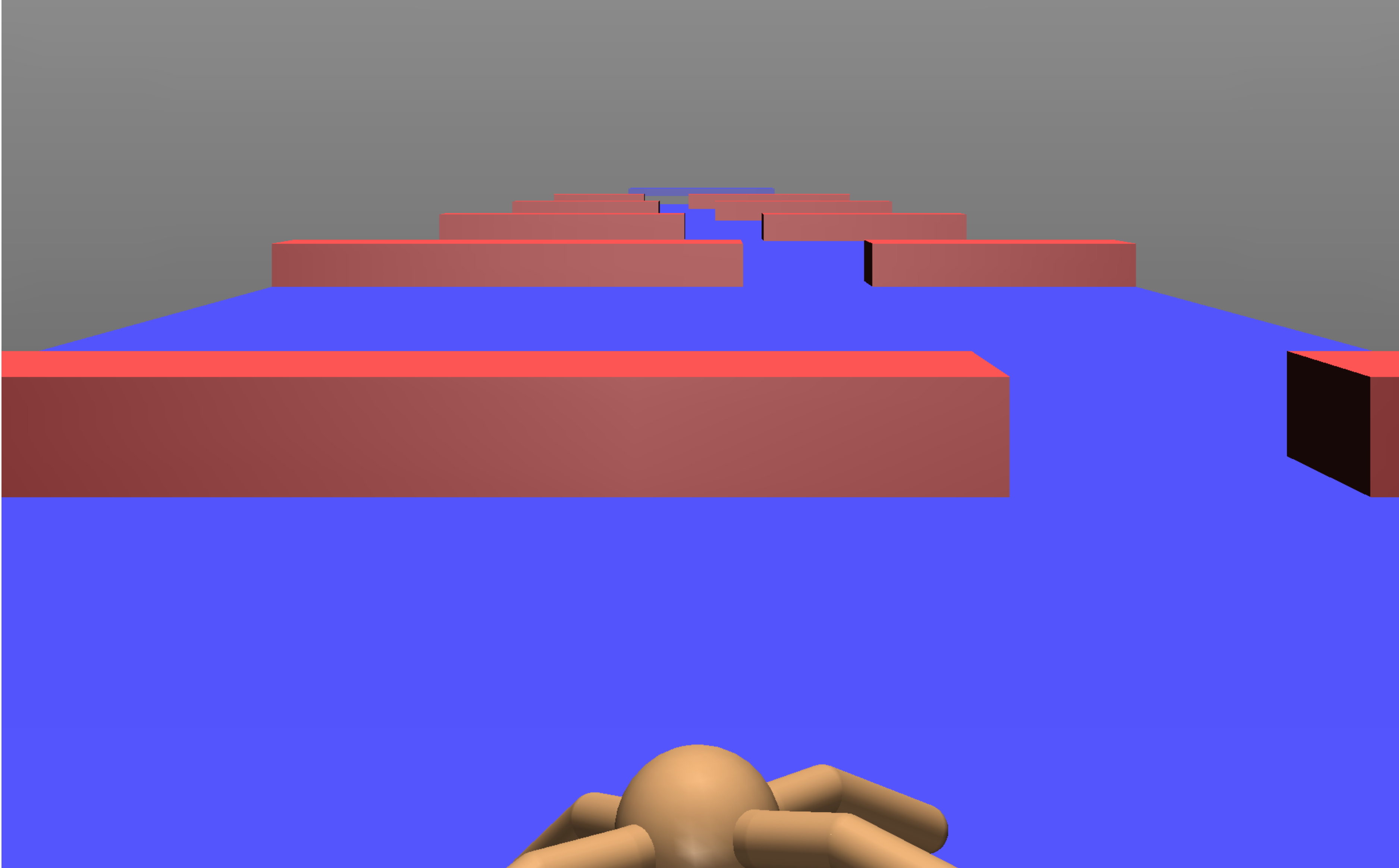}
    \includegraphics[width=0.3\linewidth]{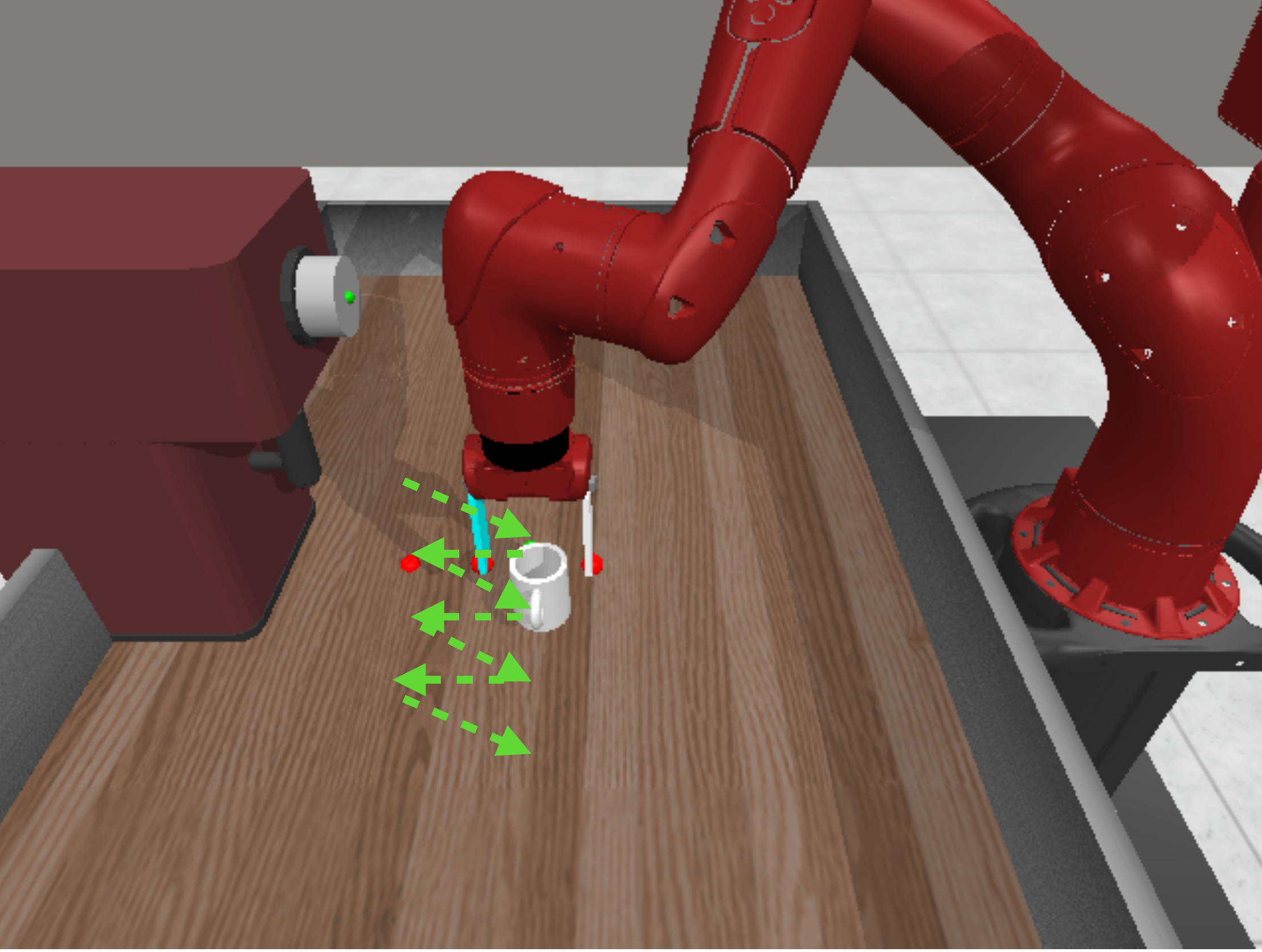} 
    \includegraphics[width=0.34\linewidth]{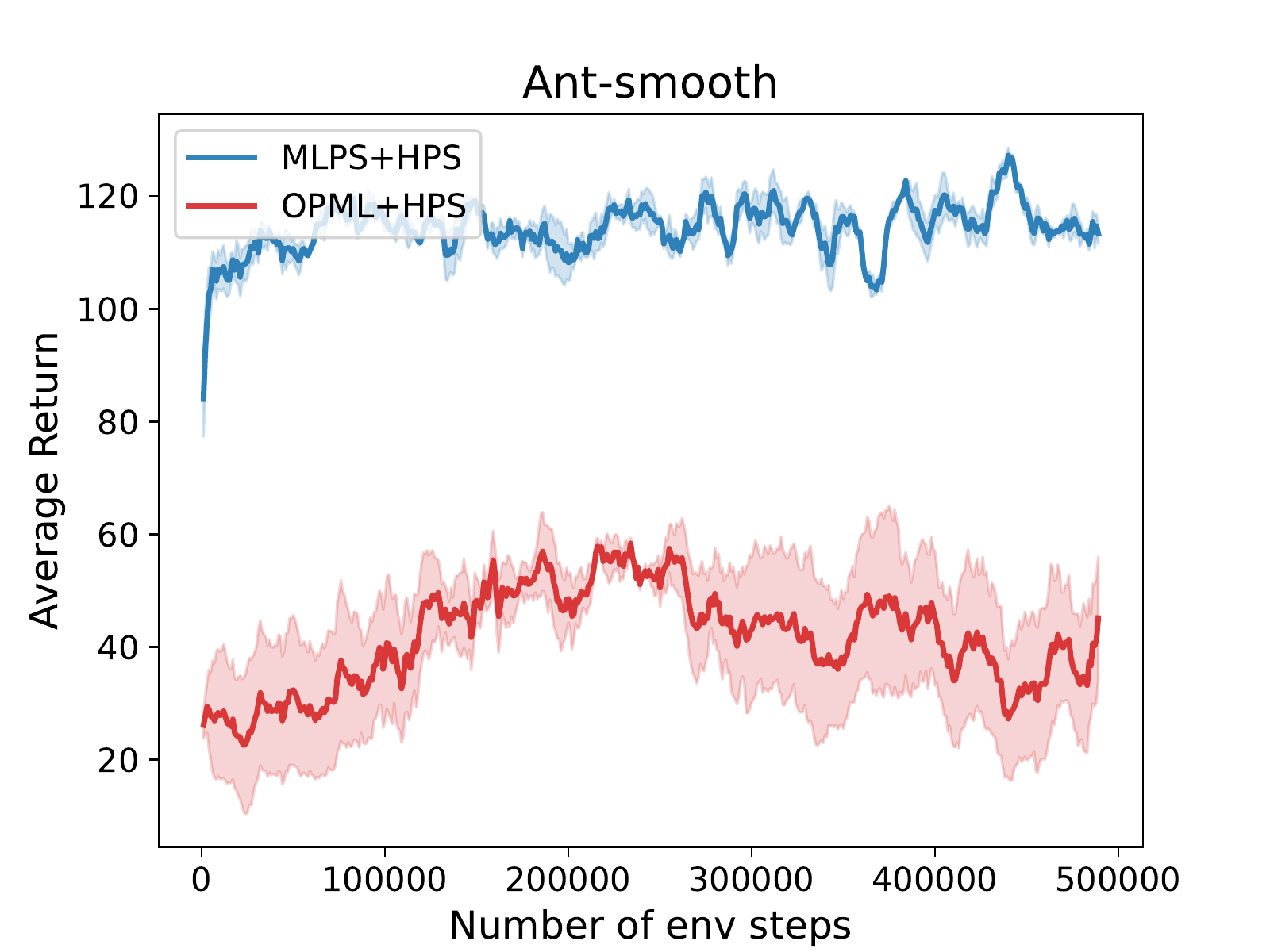}
    \includegraphics[width=0.33\linewidth]{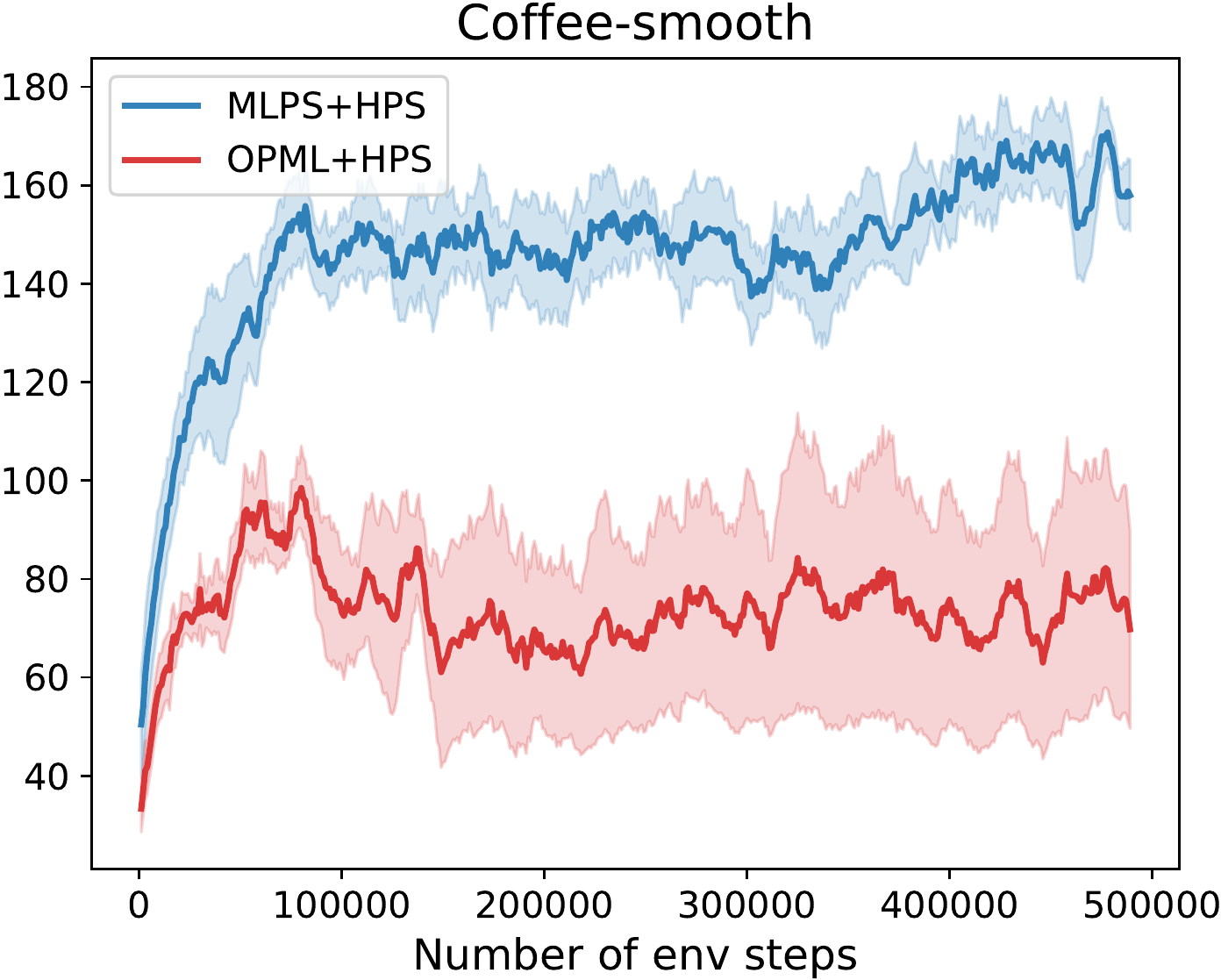}
    \caption{Comparison of the proposed MLPS algorithm with Off-Policy Meta-RL (OPML) on two scenarios where the underling continuous parameters smoothly change across subtasks.} 
    \label{fig:smoothchangeing}
\end{figure}

 For smoothly changing subtasks, it’s a special case of our method where the agent needs to consistently pick one discrete skill label as the output of the high level policy, and only needs to change the continuous parameters output from the mid-level policy. We have added experiments to illustrate our algorithm’s performance in this case. Specifically, we additionally tested our algorithm in two scenarios. In the first scenario, the ant robot needs to navigate through 5 Ant-Goal barriers, and we change the position of the doorway smoothly from $7.5$ to $3.5$ for these five Ant-Goal subtasks. In the second scenario, the robot arm needs to push and pull the mug consecutively to $12$ different locations, and we change the goal location smoothly from $[0.9, 0.57, 0]$ to $[-0.9, 0.64, 0]$. As shown in Figure~\ref{fig:smoothchangeing}, the proposed algorithm MLPS+HPS outperforms Off-policy Meta-RL(OPML) + HPS in both scenarios, indicating that our algorithm can be applied on long-horizon tasks with smooth variations between subtasks as well. For tasks where the differences between tasks changes smoothly, the high-level policy as a result only needs to output the only one discrete skill label being used and the mid-level policy will output the continuous parameters corresponding to the smoothly changing variations between subtasks.

\subsection{The difficulty of Long-horizon tasks for RL}
\label{app:longhor}
The first problem that stems from this long horizon is that a single policy neural network based on the primitive actions needs to be able to handle the distinct changes of the environment at different stages during the long execution episode (e.g., In our ant obstacle course, to reach the final goal, the ant has to move pass several gaps, obstacles, bridges), which is quite difficult.
The more insidious problem is exploration. 
Because of the long action sequence needed, uninformed exploration methods are unlikely to be successful: In the ant obstacle course, early barriers, once mastered, should be navigated so as to maximize success probability (requiring a low exploration rate), while grappling with later barriers should involve collecting enough data to pass the barrier (high exploration rate). These two problems make learning to solve such long-horizon tasks at the level of primitive actions highly difficult. 

\subsection{More discussion about the smoothness term for learning parameterized skills}
Note that in all our experiments, we use PEARL + contrastive loss as the Off-policy Meta-RL baseline. However, the proposed smoothness loss can be directly applied to other Off-policy Meta-RL algorithms as well, as it functions as an auxiliary loss to train the context encoder. Moreover, we use Dynamic Time Warping to calculate the smoothness loss and it works well for navigation tasks and robot manipulation tasks. Its effectiveness is unknown for other problems (in particular, when the observations are all in images). However, as we shown in Section~\ref{exp:smooth}, smoothness is a import factor to consider if we are trying to let an agent learn a new action space.And it will be one of the key parts to connect skill learning (MLPS) and using skills to learn (HPS) no matter what the correct form of the smoothness loss is for a specific task.

\subsection{Mathematical notation table}
\begin{table}[htb]
\centering
\caption{Mathematical Notation table}
\begin{tabular}{ll}
\centering

 Symbol & Meaning  \\\hline 
 $s$ & state \\
 $s'$ & next state \\
 $S$ & state space \\
 $A$ & primitive action space \\
 $H$ & parameterized action space \\
 $T$ & transition function\\
 $R$ & reward function\\
 $\gamma$ & discount factor\\
 $k$ & discrete action\\
 $z_k$ & continuous parameter corresponding to $k$ \\
 $K$ & total number of discrete actions \\
 $\theta$ & hidden parameter \\
 $P_{\Omega}$ & Underlining distribution over the hidden parameter \\
 $\pi$ & policy \\
 $\pi_h, \pi_{\theta_d}$ & high-level policy \\
 $\pi_m, \pi_{\theta_c}$ & mid-level policy \\
 $\pi_l$ & low-level policy \\
 $Q$ & critic \\
 $\phi$ & context encoder \\
 $\phi_{target}$ & target context encoder \\
 $\tau$ & trajectory \\
 $\kappa$ & scale of the DTW distance \\
 $\theta_a$ & parameterize the actor network when running Off-policy Meta-RL\\
 $\theta_c$ & parameterize $\pi_m$ (in the main text), parameterize the critic network (in algorithm 1)\\
 $\theta_d$ & parameterize $\pi_h$ 
 \\
 $\psi$ & parameterize $Q$ 
 \\
 $D_{KL}$ & KL divergence\\
 $\mu$ & task during MLPS training
 \\
 $b, B$ & transition batch
 \\
 $C$ & Number of tasks within one meta batch during MLPS training
 \\
 $f$ & calculate the similarity score (contrastive learning)\\
 $W_{\psi}(s)$ & Partition function (see the SAC paper)
 \\\hline

\end{tabular}

\label{tab2}
\end{table}

\end{document}